\documentclass{article}

\usepackage{microtype}
\usepackage{graphicx}
\usepackage{subcaption}
\usepackage{booktabs} 
\usepackage{csvsimple} 

\makeatletter
\csvset{
  autotabularcenter/.style={
    file=#1,
    after head=\csv@pretable\begin{tabular}{|*{\csv@columncount}{c|}}\csv@tablehead,
    table head=\hline\csvlinetotablerow\\\hline,
    late after line=\\,
    table foot=\\\hline,
    late after last line=\csv@tablefoot\end{tabular}\csv@posttable,
    command=\csvlinetotablerow},
  autobooktabularcenter/.style={
    file=#1,
    after head=\csv@pretable\begin{tabular}{*{\csv@columncount}{c}}\csv@tablehead,
    table head=\toprule\csvlinetotablerow\\\midrule,
    late after line=\\,
    table foot=\\\bottomrule,
    late after last line=\csv@tablefoot\end{tabular}\csv@posttable,
    command=\csvlinetotablerow},
}
\makeatother

\newcommand{\csvautobooktabularcenter}[2][]{\csvloop{autobooktabularcenter={#2},#1}}

\usepackage{hyperref}

\usepackage[accepted]{icml2023}

\usepackage{amsmath}
\usepackage{amssymb}
\usepackage{mathtools}
\usepackage{amsthm}

\usepackage[capitalize,noabbrev]{cleveref}

\theoremstyle{plain}

\theoremstyle{definition}

\theoremstyle{remark}

\usepackage[textsize=tiny]{todonotes}

\icmltitlerunning{Target Domain Data induces Negative Transfer in Mixed Domain Training with Disjoint Classes}

\begin{document}

\twocolumn[
\icmltitle{Target Domain Data induces Negative Transfer in \\
           Mixed Domain Training with Disjoint Classes}



\icmlsetsymbol{equal}{*}

\begin{icmlauthorlist}
\icmlauthor{Eryk Banatt}{equal,apl}
\icmlauthor{Vickram Rajendran}{equal,apl}
\icmlauthor{Liam Packer}{apl}
\end{icmlauthorlist}

\icmlaffiliation{apl}{The Johns Hopkins University Applied Physics Lab, Laurel, Maryland, United States}

\icmlcorrespondingauthor{Eryk Banatt}{ebanatt@riotgames.com}
\icmlcorrespondingauthor{Vickram Rajendran}{vickram@applied.co}
\icmlcorrespondingauthor{Liam Packer}{liam.packer@jhuapl.edu}

\icmlkeywords{Machine Learning, ICML}

\vskip 0.3in
]



\printAffiliationsAndNotice{}  

\begin{abstract}
In practical scenarios, it is often the case that the available training data within the target domain only exist for a limited number of classes, with the remaining classes only available within surrogate domains. We show that including the \textit{target domain} in training when there exist disjoint classes between the target and surrogate domains creates significant negative transfer, and causes performance to significantly decrease compared to training without the target domain at all. We hypothesize that this negative transfer is due to an intermediate shortcut that only occurs when multiple source domains are present, and provide experimental evidence that this may be the case. We show that this phenomena occurs on over 25 distinct domain shifts, both synthetic and real, and in many cases deteriorates the performance to well worse than random, even when using state-of-the-art domain adaptation methods.
\end{abstract}

\section{Introduction}

Machine Learning models have shown impressive results on several tasks in computer vision such as image classification, object detection, and semantic segmentation \citep{michalski2013machine}. However, it is still very unclear how to train models when the operational domain is significantly different from the training and validation sets. These \textbf{domain shifts} capture a large portion of machine learning failure cases and there has been growing research into the nature of these shifts and how they influence model performance \citep{zhou2021domain}.

\textbf{Category shift} \citep{xu2018deep} is a specific instance of domain shift where the class distribution in the testing domain is significantly different from that of the training domain. This sometimes manifests as class imbalance in the training domain that does not exist in the test, or in some cases, having classes in the test domain that do not exist in the training set at all. In this setting, it is common for a machine learning practitioner to attempt to alleviate this problem by augmenting their training dataset with examples of the rare classes, even if that data comes from different domains. For example, they may use data from KITTI \citep{Geiger2012CVPR} to augment the rare class of bicycles in NuImages \citep{nuscenes2019}, as both have similar taxonomies and tasks. This intuitively helps give the model some examples of features of the rare or missing classes. 
\begin{figure}
    \centering
    \includegraphics[width=0.75\linewidth]{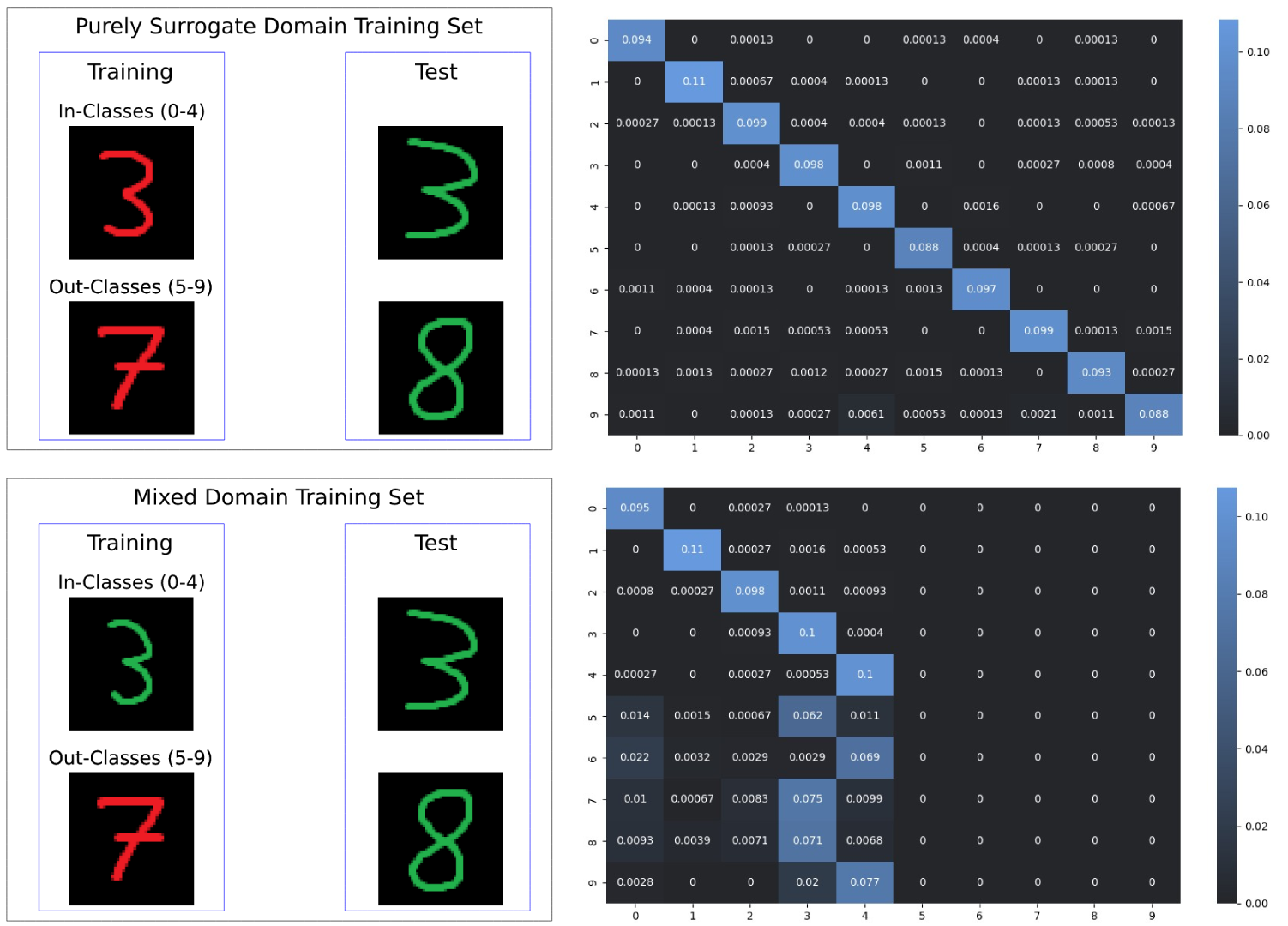}
    \caption{Colored MNIST example of the negative transfer that we study. Note that adding target domain (green) data to the training set significantly reduces performance on the target dataset.}
    \label{fig:mnist_data}
\end{figure}

In scenarios where there are none or very few examples of the missing classes in the training set \citep{wang2020generalizing}, augmenting those classes with examples from other datasets will sometimes appear to improve performance on those rare classes. Intuitively, if the newly-introduced domain is "not too far away" from the testing domain, it is possible to train a classifier using that data which will generalize to unseen points in the target domain \citep{beery2020synthetic}. However, we find that this may be a misleading comparison: rather than comparing to few-shot or low-shot learning settings, where performance on these classes is typically not possible, and any performance constitutes an improvement, the appropriate benchmarks are the settings where all classes are trained on either target domain data or surrogate domain data, rather than a class-wise mix between the two.

We perform simple experiments to demonstrate a counterintuitive phenomenon about this mixed-domain training paradigm -- under category shift, models that are trained \textbf{solely} on the "out-of-domain" data perform significantly better compared to models that are trained on both the "out-of-domain" data as well as training data from the domain of interest. This is especially surprising for several reasons: first, we are simply adding data from the domain of interest, which intuitively should bring our new training set "closer" to the domain of interest; second, the data that we are adding consists of classes that are completely disjoint from the "out-of-domain" data that we have added, and so we would not expect it to harm the generalization of the "out-of-domain" classes. Within the transfer learning literature, this can be thought of as the reverse of the common bottleneck of Negative Transfer (NT) \citep{zhang2020survey}, where adding "dissimilar" source data will damage performance on the target data.

Figure \ref{fig:domain_shift} shows how this setting relates to other common settings when studying domain shift. To the best of our knowledge, there are no other existing works that study this inverse problem of negative transfer induced by adding target domain data in other classes. This finding is especially relevant since it is extremely likely that these effects occur on existing datasets that do not have explicitly defined domains; for example, if every picture of a single class in CIFAR10 was collected at night, or by a different camera, then we would expect this negative transfer to occur.

Our contributions are threefold - First, we use a toy example to identify a previously unknown source of negative transfer in the common practical scenario of training in a mixed-domain setting with some classes in the target domain and some classes in a surrogate domain. Second, we analyze the nature of this negative transfer and empirically show that it is induced by adding data from the \textbf{target} domain, and that adding data from other unrelated domains does not exhibit this same negative transfer. Finally, we show that these behaviors occur on over 25 domain shifts in real image classification datasets such as PACS, VLCS, and VisDA2017, and that \textbf{even the presence of state-of-the-art domain adaptation algorithms} does not significantly address this problem. Our code is available in the supplemental material.  

\section{Definitions and Notation}

We borrow notation from the multi-source domain adaptation literature and use $\mathcal{D} = \{\mathcal{D}_1, \mathcal{D}_2, \cdots,\mathcal{D}_n\}$ to denote a set of $n$ domains. We define one of these domains to be $\mathcal{T} = \mathcal{D}_t$, the \textbf{target domain}, and another of these domains $\mathcal{S} = \mathcal{D}_s$ to be the \textbf{surrogate domain}. Note that we use the term \textit{surrogate} in place of \textit{source} to help clarify that we are not interested in methods that improve the domain adaptation strictly from $\mathcal{S}$ to $\mathcal{T}$, but instead are interested in using $\mathcal{S}$ as a replacement for missing or rare classes in $\mathcal{T}$. 

Each $\mathcal{D}_i$ contains $k_i$ classes $\mathcal{C}_i = \{c_{i_1}, c_{i_2}, \cdots, c_{i_{k_i}} \}$, and we define $\mathcal{C}$ as the union of the $\mathcal{C}_i$. For simplicity, we will assume that all domains have the same shared taxonomy, and so each of these $\mathcal{C}_i$ are the same $\mathcal{C}$. We split this $\mathcal{C}$ into two disjoint subsets, $\mathcal{C}_I$ and $\mathcal{C}_O$, and denote these subsets as \textbf{in-classes} and \textbf{out-classes} respectively.

Let $\mathcal{T}_{train}, \mathcal{T}_{test}$ be drawn i.i.d from $\mathcal{T}$. We split $\mathcal{T}_{train}$ into $\mathcal{T}_{in|train}, \mathcal{T}_{out|train}$ where $\mathcal{T}_{in|train}$ is the subset of $\mathcal{T}_{train}$ that has labels in $\mathcal{C}_I$, and similarly $\mathcal{T}_{out|train}$ to be the subset with labels in $\mathcal{C}_O$, and construct $\mathcal{T}_{in|test}, \mathcal{T}_{out|test}$ in the same way. We perform the same procedure on $\mathcal{S}$ to create subsets of the training and testing domains that are restricted to the in-classes and out-classes.

\begin{figure}
    \centering
    \includegraphics[width=1\linewidth]{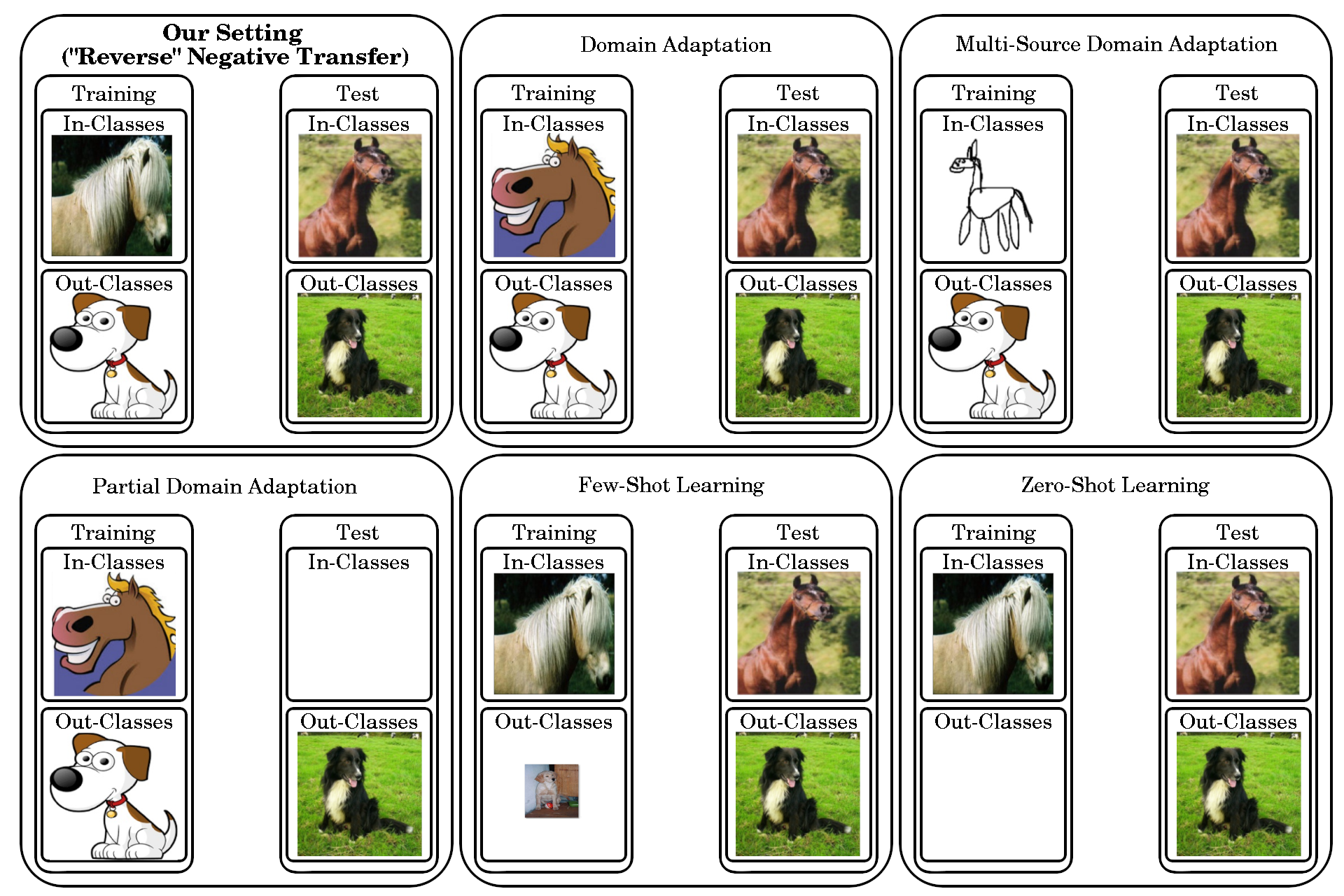}
    
    \caption{The settings related to our work. We are specifically interested in how performance in our setting differs from simply performing supervised learning in the domain adaptation setting.}
    \label{fig:domain_shift}

\end{figure}

With this notation, we can now clearly state the problem setting: We are specifically interested in the performance of models trained on $\mathcal{T}_{in|train} \cup \mathcal{S}_{out|train}$ tested on $\mathcal{T}_{out|test}$, compared to models trained on $\mathcal{S}_{out|train}$ on the same testing set $\mathcal{T}_{out|test}$ - Figure \ref{fig:domain_shift} shows that this is essentially a comparison between extremely naive domain adaptation from $\mathcal{S}_{out}$ to $\mathcal{T}_{out}$ and this mixed-domain setting where we simply add $\mathcal{T}_{in}$ to the training set. We find that this setting is extremely underexplored in the literature, and that models trained on the mixed-domain dataset surprisingly perform much worse than models trained solely on the surrogate domain, even though they intuitively do not seem like they would be harmed by adding target domain data from disjoint classes. 

\section{Related Work}

\subsection{Missing or Rare Classes}

Class imbalance is well known to damage performance in machine learning models by biasing performance towards the more common classes \citep{buda2017systematic}. The three primary strategies for dealing with class imbalance are 1) over-sampling minority data \citep{mohammed2020sampling} \citep{yap2014application}, 2) under-sampling majority data \citep{mohammed2020sampling} \citep{liu2008exploratory} \citep{yap2014application}, and 3) augmenting the rare classes with examples from other datasets \citep{beery2020synthetic} \citep{chawla2002smote} \citep{he2008adasyn} \citep{lanzini2021image}. Our work focuses on the third approach, which appears to avoid the information loss of under-sampling as well as the overfitting risk of over-sampling by using a different dataset to augment the rare or missing classes.

\citet{beery2020synthetic} show a constrained scenario where rare classes are augmented with synthetic examples in order to improve performance on those rare classes. In that work, the authors saw the best performance by augmenting their long-tailed dataset with synthetic data generated via a mix of pasting real segmented animals onto real backgrounds, and generating images in Unity \citep{haas2014history}. The authors note that despite aiding performance in their dataset, the features generated by synthetic points appear to be highly dissimilar from the same class within the target domain, clustering entirely apart. \citet{das2021domain} improved upon this by using domain adaptation techniques during training, but still found that despite performance improvements the features seemed to cluster apart. Likewise, \citet{konushin2021road} trained a model to add synthetic road signs to images, in order to improve performance on rare signs both for classification and detection. In both cases, performance of these models was evaluated by comparing low-shot training to mixed-domain training, showing an improvement under those settings.

\subsection{Domain Adaptation}

A variety of works exist which seek to learn from one or multiple non-target domains, with varying degrees of success \citep{sun2015survey} \citep{wang2019towards} \citep{rebuffi2018efficient} \citep{rebuffi2017learning}. In these works, the focus is on an established shift between some domain (or group of domains) and another, distinct domain; more specifically, on techniques which can be used to overcome that drop in performance. In contrast, our work posits a special case of multi-source domain adaptation under category shift, where the set of domains used in training contains the target domain, and where the set of labels in each domain is partially or entirely non-overlapping. The closest to our setting is \citet{xu2018deep}, which explores multi-source domain adaptation with category shifts between source domains; however, their target domain is still explicitly distinct from the source domains (i.e. unseen in training). While many domain adaptation techniques could certainly be applied in the setting where we treat the target domain as one of the sources \citep{das2021domain}, generally speaking, domain adaptation is focused on techniques to \textbf{improve} performance from the source to the target; in contrast, our focus is on how additional target domain data in the training set \textbf{decreases} performance in specific scenarios.

Open-set recognition \citep{mahdavi2021survey} and open-set domain adaptation \citep{jain2014multi} \citep{liu2019separate} \citep{panareda2017open} similarly consider settings where the test domain has classes that are not seen in the traditional training paradigm. In these settings, the goal is to assign the unseen classes to an "unknown" class, whereas in our setting we attempt to assign the correct class based on augmenting the training set with out-of-domain examples of the unseen classes. \citep{das2021domain} \citep{beery2020synthetic} \citep{konushin2021road}.

A slightly closer related family of techniques is that of partial domain adaptation \citep{cao2018partial} \citep{li2020deep} \citep{gong2021mdalu}. In this setting, we seek to transfer from a source domain which has a superset of the classes as is needed for the target domain. The main distinction between our setting and partial domain adaptation is that the source and target domains are distinct, in our setting they are only distinct on a class-conditional basis: some classes in the training setting are from the source domain, and others are from the target domain, which is a setting we propose is paradoxically harder rather than easier. However, an important similarity between our setting and partial domain adaptation is the "outlier source classes" causing negative transfer \citep{zhang2020survey}, which is observed in our setting for "outlier target classes".

\section{Motivating Example: ColoredMNIST}

We begin with a motivating toy example to demonstrate this effect. We randomly split the MNIST \citep{lecun1998mnist} dataset into two equal subsets, $\mathcal{R}$ and $\mathcal{G}$, and color the digits red and green respectively. We use the digits $0$-$4$ to be the in-classes and $5$-$9$ to be the out-classes, and construct the splits $\mathcal{G}_{in|train}, \mathcal{G}_{out|train}, \mathcal{G}_{in|test}, \mathcal{G}_{out|test}$ as defined in Section 2, and do the same for $\mathcal{R}$. We set $\mathcal{G}$ to be the target domain, and $\mathcal{R}$ to be the surrogate. We train a simple neural network (experimental details in supplemental material) under various training and testing splits, and examine its performance. Note that all of these models train with a full ten class label space, even if their training data only contains data with support for a subset of those labels. 

We show the results of these experiments in Table \ref{table:colored_mnist}. This domain shift is small enough that simply training on $\mathcal{R}_{out|train}$ adapts reasonably well to $\mathcal{G}_{out|test}$. Interestingly, we see that there is a significant difference between the performance on the in-classes and the out-classes in this mixed domain setting, and it is exactly the out-classes that are causing the reduced performance of the mixed-domain model. This shows a remarkable negative transfer that occurs simply by adding $\mathcal{G}_{in|train}$ into the training set, which completely deteriorates all domain generalization from training on $\mathcal{R}_{out|train}$ and testing upon $\mathcal{G}_{out|test}$. Note that while the overarching test set performance decreases dramatically, the performance on the in-classes does not change at all compared to when the model trained only on data from $\mathcal{G}_{train}$. Given this, we focus the remainder of our experiments purely on studying the performance of the out-classes in the test domain.

\begin{figure}
    \centering
  \includegraphics[width=1\linewidth]{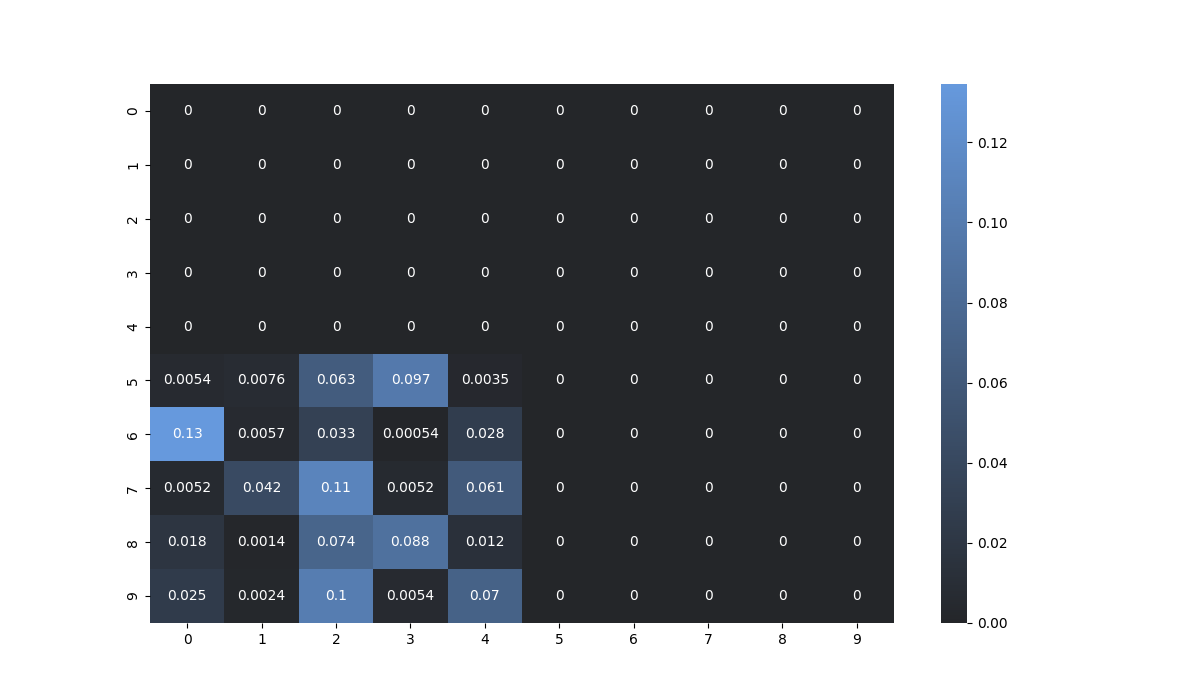}%
  \caption{A confusion matrix for a model trained on green digits 0-4 and red digits 5-9, tested upon green digits 5-9. In each case, the model predicts that a test digit must be one of the classes 0-4, and picks pseudorandomly.}%
  \label{fig:conf_mnist}
\end{figure}
\begin{table}  
 \caption{Negative Transfer in Colored MNIST. Performance on out-classes degrades dramatically when target data is added to the training set, even though only the added target data is of disjoint classes.}%
  \label{table:colored_mnist}
  \centering
  \begin{tabular}{lcc}
    \cmidrule(r){1-3}
    Training & Test     & Accuracy \\
    \midrule
    $\mathcal{G}_{train}$ & $\mathcal{G}_{test}$ & 0.968   \\
    $\mathcal{R}_{train}$ & $\mathcal{G}_{test}$ & 0.962       \\
    $\mathcal{R}_{out|train} \cup \mathcal{G}_{in|train}$ & $\mathcal{G}_{test}$ & 0.498   \\
    $\mathcal{G}_{out|train}$ & $\mathcal{G}_{out|test}$ & 0.974    \\
    $\mathcal{R}_{out|train}$ & $\mathcal{G}_{out|test}$ & 0.972       \\
    $\mathcal{R}_{out|train} \cup \mathcal{G}_{in|train}$ & $\mathcal{G}_{out|test}$ & 0.000   \\
    $\mathcal{R}_{out|train} \cup \mathcal{G}_{in|train}$ & $\mathcal{G}_{in|test}$ & 0.981   \\
    \bottomrule
  \end{tabular}
 
\end{table}

\begin{figure}
    \centering
    \includegraphics[width=0.5\linewidth]{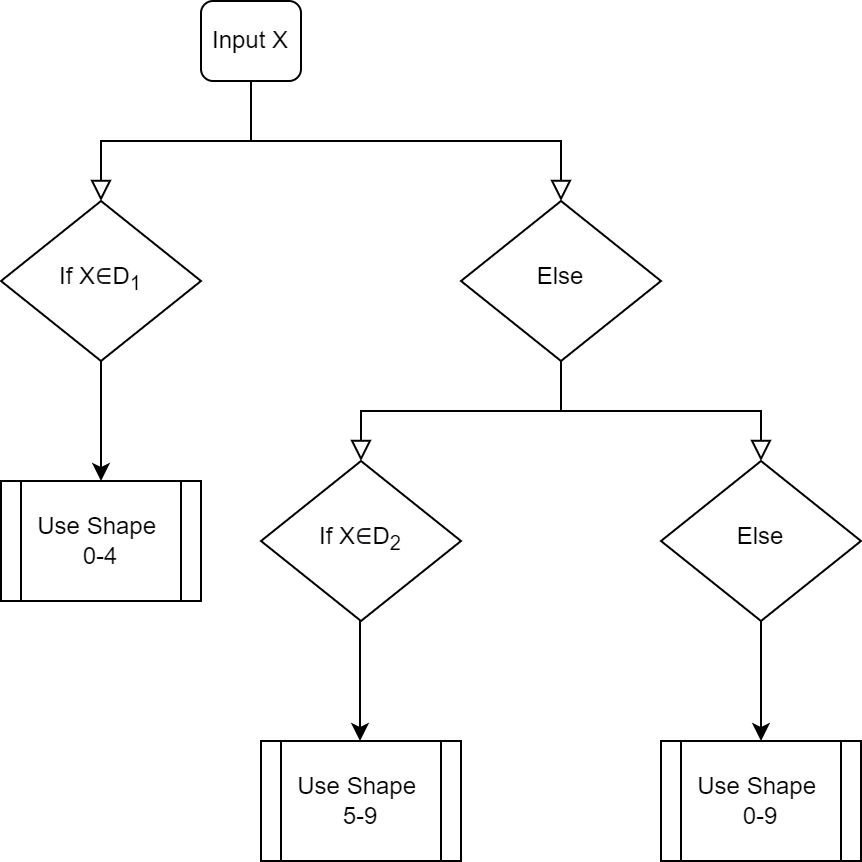}
    
    \caption{The apparent decision making process of our model. We show that a loosely predictive domain label will prevent the model from using a more predictive feature it has learned before, if that domain is a domain seen in training for a class-domain combination which was previously unseen.}
    \label{fig:dec_tree}
\end{figure}



We show the confusion matrix of the model trained on $\mathcal{R}_{out|train} \cup \mathcal{G}_{in|train}$ evaluated on $\mathcal{G}_{out|test}$ in Figure \ref{fig:conf_mnist}. Now things become much more clear - this model is effectively classifying a green digit from $5$-$9$ randomly between one of the classes $0$-$4$, and is now completely ignoring any shape information it may have gained by training on the red digits from $5$ to $9$. 




We note the similarity of this experiment to the shiftMNIST experiment in \citep{jacobsen2018excessive}, wherein each class of MNIST has a location-based feature which perfectly encodes the class label. However, in that work the number of "shifts" was equal to the number of classes, and when the encoding feature was removed the model could not reliably infer upon the points due to not having learned the relevant discriminative feature. We claim the model \textit{has} learned the relevant discriminative feature in our setting, as learning it is necessary to tell apart the classes within the in-classes and out-classes, as simple domain discrimination is not enough to solve this task.

We would expect that the model trained on mixed-domain data would infer upon the signal from the shape, which is more predictive of class, in order to tell apart the classes. But what we instead find is that the network learns a "lazy" rule first \citep{geirhos2020shortcut} \citep{ilyas2019adversarial}, where it eliminates specific classes based on the digit's color, and then selects the predicted class based on the shape of the digit. This means that the model that trains on both target domain data and surrogate data effectively splits the domains first, and then decides which class to predict depending on the domain that it lives in. We depict this process in Figure \ref{fig:dec_tree}.

This is the negative transfer effect which arises from mixed-domain training; adding examples from the target domain for in-classes  will damage performance on points in the target domain for the out-classes, even if the training data used for those points would, by themselves, be sufficient for generalizing to the target domain.

\section{Experiments}

We perform a series of experiments to show that our findings on Colored MNIST generalize to more complex datasets and models. 

We use the following datasets: 
\begin{itemize}
    \item \textbf{PACS} \citep{li2017deeper} is a dataset traditionally used for domain generalization. Each of the four domains (Photo, Art Painting, Cartoon, and Sketch) consist of seven classes ranging from houses to people. For all experiments in this setting, we use the first three classes to be our in-classes, and the remaining four to be our out-classes. We perform experiments on each pairwise domain shift, which yields a total of 12 domain shifts from this dataset. PACS is an open-source dataset not released under any license.
    
    \item \textbf{VLCS} \citep{fang2013unbiased} is another dataset traditionally used for domain generalization. Each of the four domains (VOC2007, LabelMe, CalTech101, and SUN09) consist of five classes ranging from birds to people. For all experiments in this setting, we use the first two classes to be our in-classes, and the remaining three to be our out-classes. We perform experiments on each pairwise domain shift, which yields a total of 12 domain shifts from this dataset. VLCS is an open-source dataset not released under any license.
    
    \item \textbf{VisDA2017} \citep{fang2013unbiased} is a dataset of 12 classes traditionally used for domain adaptation between its two domains of synthetic data and real data. The synthetic data is generated using CAD models while the real data is collected from a  photorealistic validation domain. We do not use the testing data downloaded from VisDA, and only consider the domain shift from the synthetic training domain to the real validation domain as our two domains. In these experiments, we use the first 6 classes as the in-classes. We perform experiments on each pairwise domain shift, which yields a total of 2 domain shifts from this dataset. VisDA2017 was released under the MIT license.
\end{itemize}

For all experiments, we run five trials with different random seeds on training and testing splits. Each dataset is divided into 80\%, 10\%, 10\% between training, validation, and testing as described in Section 2. For consistency, all results in this section are completed with a ResNet18 model that has been pretrained on Imagenet; results on other models including ViT-B, VGG, and ResNet50's can be found in the supplemental material. All experiments take place on a single NVIDIA 1080TI, and code to reproduce these experiments is contained in the supplemental material. A more complete description of the implementation details and hyperparameters are also available in the supplemental material. 

\subsection{Negative Transfer vs. Simple Domain Shift}

\begin{figure}
    \centering
    \includegraphics[width=\linewidth]{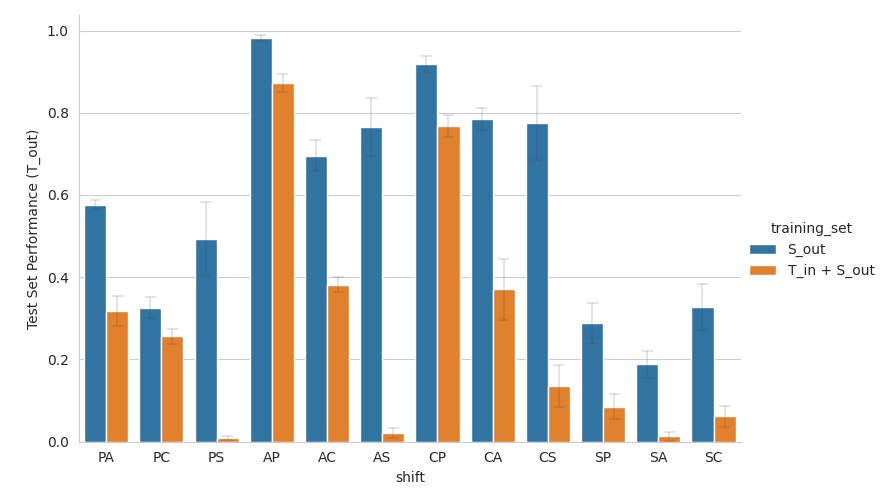}
    \caption{Performance of PACS models with and without target domain in-classes added to the training set. We use the word "shift" to denote the target and surrogate domains, e.g. a shift of "PA" means $\mathcal{S}$ = P and $\mathcal{T}$ = A. In virtually every case, performance on target domain out-classes ($\mathcal{T}_{out}$) is much worse with the addition of target domain data for in-classes.}
    \label{fig:pacs_shift}
\end{figure}

 Tables \ref{table:pacs_resnet} and \ref{table:vlcs_resnet} summarize the results of these experiments. Each of these tables show the performance of a model trained purely on $\mathcal{S}_{out|train}$ compared to a model trained on $\mathcal{S}_{out|train} \cup \mathcal{T}_{in|train}$, on the test set of $\mathcal{T}_{out|train}$. As we can see from Figure \ref{fig:pacs_shift}, the performance of the mixed domain model is significantly worse on almost every domain shift. For some of the domain shifts, particularly those involving Sketch, we see that adding the target domain brings the performance of the models to far worse than random guessing.

\begin{figure*}
    \centering
    \includegraphics[width=1\linewidth]{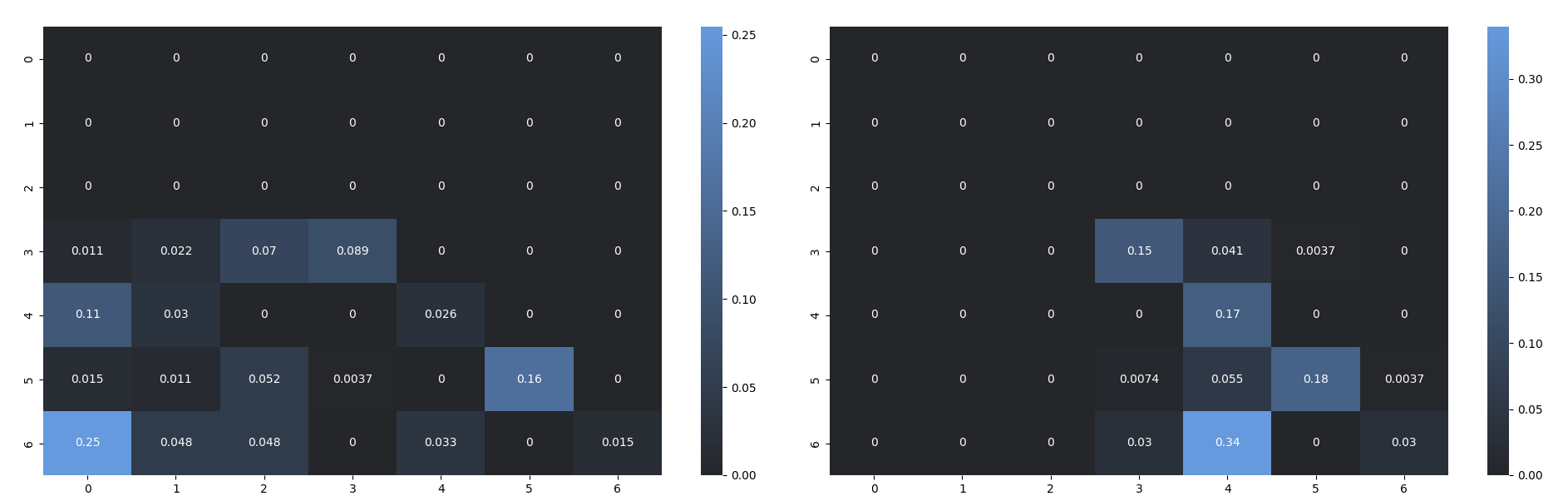}
    \caption{PACS Confusion Matrix for A out-classes with target domain A and surrogate domain P with (left) and without (right) the addition of in-classes from the target domain. Note the diagonal like structure in the right figure that disappears in the left figure. When target domain data for in-classes is added to training, the model will often pseudorandomly guess one of the in-classes whenever it tests on a point in the target out-classes.}
    \label{fig:pacs_conf}
\end{figure*}

 The confusion matrices in Figure \ref{fig:pacs_conf} also show that the same phenomena occurs that we saw in the Colored MNIST case - when target domain data is added into the training set, new examples of target domain data, even if it is of a completely different class, get randomly assigned to the in-classes. However, the model that trains without any target domain data has a fairly diagonal confusion matrix in the out-classes, with much fewer misclassifications to the in-classes.

\begin{table}
\centering
\caption{Negative Transfer on PACS, where the test set is $\mathcal{T}_{out|test}$ for each target domain.} 
\label{table:pacs_resnet}
\csvautobooktabularcenter[]{pacstable.csv}
\end{table}

\begin{table}
\centering
\caption{Negative Transfer on VLCS, where the test set is $\mathcal{T}_{out|test}$ for each target domain.} 
\label{table:vlcs_resnet}
\csvautobooktabularcenter[]{vlcstable.csv}
\end{table}


\subsection{Multi-Source Domain Shift}


In this experiment, we show that the negative transfer is strictly a consequence of adding target data, rather than simply the addition of another domain.

\begin{figure}
    \centering
    \includegraphics[width=0.9\linewidth]{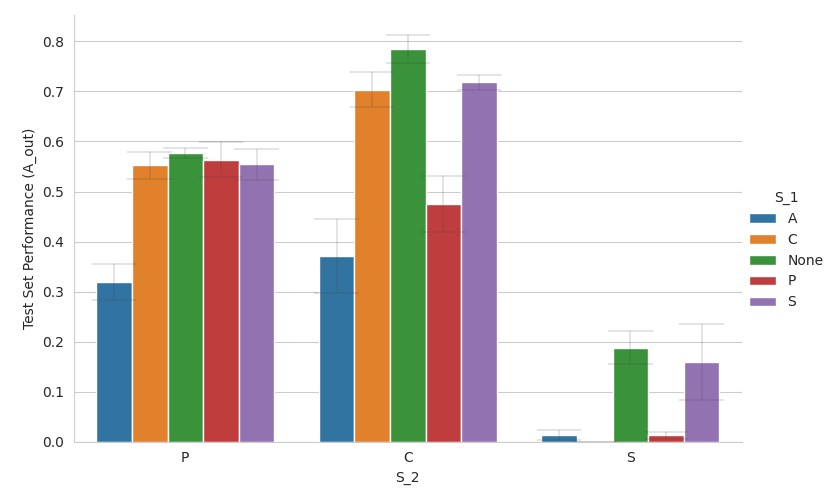}
    \caption{Accuracy of PACS models trained with $\mathcal{S}_{1|in|train} \cup \mathcal{S}_{2|out|train}$, where the target domain is Art (A). Note that when $\mathcal{S}_1 = A$, there is significantly worse performance compared to the other settings.}
    \label{fig:pacs_s2_bar}
\end{figure}

We define $\mathcal{S}_1, \mathcal{S}_2$ to be two surrogate datasets, and train on $\mathcal{S}_{1|in|train} \cup \mathcal{S}_{2|out|train}$. We evaluate these models on the same test sets as before $\mathcal{T}_{out|test}$, to see whether it is specifically the case that using $\mathcal{T}$ in place of a different domain $\mathcal{S}_1$ causes the negative transfer. Figure \ref{fig:pacs_s2_bar} shows the result of this experiment on the PACS dataset, when $A$ is the target domain. We can clearly see that when $A = \mathcal{S}_1$, the performance is significantly worse compared to when $\mathcal{S}_1$ is any other dataset. In fact, the performance of the non-target domain datasets is very similar to the performance of $\mathcal{S}_1 = \emptyset$, which shows that adding those domains as in-classes to the training data does \textbf{not} induce the same negative transfer we see from the target domain. 

Interestingly, some surrogate domains do exhibit some negative transfer, specifically in the cases when $S_1 = P$, $S_2 = \{C,S\}$, $T = A$ in Figure \ref{fig:pacs_s2_bar}). This makes some intuitive sense since the Photo domain looks visually similar to Art, however, further analysis on the relationship between the degree of negative transfer and non-target domains is outside the scope of this paper. Full results on this experiment and graphs for the remaining domains are contained in the supplemental material.

We bolster this claim by examining what happens when the in-class and out-class domain distribution is more similar - in the supplemental material, we train on $\mathcal{S}_{in|train} \cup \mathcal{T}_{in|train} \cup \mathcal{S}_{out|train}$, and continue to test on $\mathcal{T}_{out|test}$ and show that the negative transfer still has a significant presence, similar to that of having no surrogate data in the in-classes.

\subsection{Proportion of Target Domain Data}
So far, we have shown that negative transfer is induced primarily when the target and surrogate domains have no class overlap in the training set. However, it is clear that adding some proportion $p$ of $\mathcal{T}_{out|train}$ to the training set will eventually reduce the negative transfer - in the extreme case, when $p=1$ we reduce to a supervised learning problem under no domain shift. To study how negative transfer evolves in this scenario, we perform an experiment adding different quantities of the $\mathcal{T}_{out|train}$ into the training set.
\begin{figure}
    \centering

    \includegraphics[width=\linewidth]{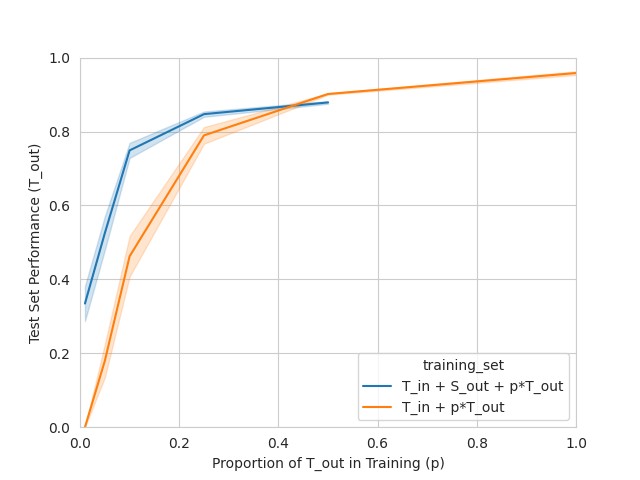}
    \caption{Performance on out-classes for a network whose training data contains some proportion of surrogate and target domain data. We observe that surrogate data helps performance for low class count, that small amounts of target domain data help performance substantially, and that greater quantities of target domain data diminish the value of surrogate data for those classes.}
  \label{fig:pacs_prop}
\end{figure}

Figure \ref{fig:pacs_prop} shows an example of these results on the PACS dataset. We find that with very low amounts of target domain data for these classes, negative transfer is still present. We find that with very low amounts of target domain data for these classes, negative transfer is still present. However, typically with a proportion of about 5-20\% of target domain  to surrogate domain, the negative transfer begins to gradually diminish - we can assume at this point that the machine learning model has enough examples of the out classes in the target domain that it no longer primarily classifies out-classes as in-class purely due to the domain information. Full results and figures for this experiment are in the supplemental material. 
\subsection{Domain Generalization Methods}
\begin{table*}
\centering
   \caption{Negative transfer between ResNet-18 models trained on $S_{out|train}$ and $S_{out|train} + T_{in|train}$ and evaluated on $T_{out|test}$ on PACS.}
     \label{table:dg_table}
\begin{tabular}{ccccccccc}
   
    \cmidrule(r){1-9}
    Shift & ERM & RSC & MixUp  & CutMix  & SD & IRM  & CDANN  & GroupDRO \\
    \midrule
    PA & 25.36 & 33.63 & 20.37 & 32.29 & 35.71 & 29.19 & 12.43 & 7.17  \\
    PC & 8.42 & 25.74 &  15.01 & 13.52 & 3.88 & 15.84 & 19.65 & 2.03  \\
    PS & 52.63 & 45.12 & 45.67 & 45.13 & 67.52 & 46.79 & 50.21 & 35.97 \\
    AP & 7.58 & 5.67 & 7.38 & 6.10 & 9.88 & 11.28 & 9.75 & 8.51  \\
    AC & 32.91 & 22.38 & 34.61 & 34.06 & 14.18 & 35.92 & 33.19 & 11.28 \\
    AS & 75.22 & 68.48 & 52.55 & 52.63 & 54.71 & 76.67 & 38.02 & 22.74 \\
    CP & 19.21 & 27.52 & 19.32 & 42.68 & 14.27 & 45.80 & 11.55 & 7.55  \\
    CA & 49.99 & 38.79 & 37.36 & 53.22 & 41.73 & 58.88 & 32.39 & 12.81 \\
    CS & 57.66 & 36.71 & 48.53 & 42.87 & 69.93 & 65.05 & 45.18 & 10.25 \\
    SP & 13.64 & 11.76 & 7.84 & 10.97 & 18.10 & 17.01 & 4.05 & 2.12  \\
    SA & 18.67 & 15.18 & 14.75 & 17.25 & 15.38 & 20.89 & 14.87 & 14.61 \\
    SC & 18.79 & 21.97 & 12.87 & 17.14 & 11.97 & 25.73 & 2.64 & 12.29 \\
    \bottomrule
  \end{tabular}

  \end{table*}
  
  We assess how much negative transfer occurs under 7 different domain generalization methods - MixUp/CutMix \cite{mixup} \cite{zhang2017mixup} \cite{cutmix}, Representation Self-Challenging (RSC) \cite{rsc} , Spectral Decoupling (SD) \cite{sd} , Group Distributionally Robust Optimization (GroupDRO) \cite{groupdro}, Invariant Risk Minimization (IRM) \cite{irm}, and Conditional Domain Adversarial Neural Network (CDANN) \cite{cdann}. Note that the first four methods (Mixup/CutMix, RSC, SD) use no domain knowledge - these methods could be applied even when the practitioner is unaware that this negative transfer is occurring. The remaining three methods (GroupDRO, IRM, CDANN) all treat the in-classes and the out-classes as separate domains, and are given explicit supervision about what the domain is. Note that in many cases, we may not explicitly know that classes are from different domains (e.g. if there was some difference in the data collection strategy for some classes); however, for simplicity we assume we have this knowledge.  

The experimental setup is the same - we train on in-classes of domain B, out-classes of domain A, and then test on out-classes of domain B. We compare the performance to just training on the out-classes of domain A and testing on the out-classes of domain B. We find in Table \ref{table:dg_table} that all of these methods fail to mitigate the negative transfer, and in some cases (IRM) perform worse than using no domain generalization at all. Even the methods that have explicit domain knowledge, and attempt to create domain-invariant features, fail to perform well in this setting. We note that GroupDRO does perform the best at mitigating the drop in performance, however, there still exists a significant negative transfer for several shifts.

\section{Discussion}

In this work we demonstrate a general phenomenon which emerges in commonly used mixed-domain training paradigms for machine learning models. Our experiments indicate that this effect is both pervasive and extremely damaging, often degrading performance to far worse than random for the classes of interest. In addition, we empirically demonstrate that the performance degradation is directly caused by the training data containing data from the target domain, and that adding a different out-of-domain set of data to the training data will improve performance compared to adding data from the target domain.

The implications of this work are clear - since resolving low class count through dataset augmentation is such a common strategy among industry machine learning practitioners, understanding the potential harms of such a strategy is of critical importance. It is also critical to note that this performance degradation can "hide" inside of experimental designs which instead make this common strategy appear to help. In the low-shot setting, for example, adding out-of-domain data to the training set could appear to improve the performance of the classifier since inferring upon these points would be essentially impossible with normal training techniques. However, this "improved" performance could actually be \textbf{degraded} performance compared to training a second model solely on that out-of-domain data for those classes of interest, even with no domain adaptation techniques whatsoever.

More broadly speaking, these insights are likely very relevant towards understanding unconscious bias inherent in classifiers as well as conducting proper dataset curation. While the experiments in this paper focus on settings where we know what each domain is, and exactly what each class is, it is very possible that there exist unknown factors of variation that are only present in some subgroups of the data outside of class. This would effectively create our mixed-domain scenario in a natural setting, and the models could be unconsciously biased towards those induced in-subgroups. Further, the effect of the negative transfer we have shown might be completely invisible, since we would not be able to separate out those subgroups. From a data collection perspective, it is very common to merge datasets together (especially when they have the same taxonomy), before releasing the data - even something as simple as data collected over two different days might exhibit this effect. 

An interesting observation is that the degree of negative transfer varies between domains, and appears to be more severe when domains are intuitively "further apart". It's possible that the propensity of a model to learn to maladaptively discriminate between domains in the presence of clear predictive features could spark further discussion towards the quantification of distance between two domains. Future work could include studying the causes of these shifts, as well as how they change when strategies beyond standard empirical risk minimization are applied.

While we have shown strong evidence for negative transfer in classification, there are several limitations of our work. First, our experiments have shown this purely in image classification; determining whether this effect occurs in object detection, segmentation, or other machine learning tasks is outside the scope of this work. Second, while we have presented an adequate strategy for mitigating the effects of this kind of negative transfer (GroupDRO), we note that this method does not fully solve this problem, and worse, requires full domain knowledge that this negative transfer is happening. Finally, while we have tested that this effect occurs on multiple different architectures and datasets, we did not have the compute to test whether this happens in very large models. 

Future work will focus on learning more powerful mitigation strategies to abate the presence of this target domain induced negative transfer. While the purpose of this work was to outline a scenario where a training distribution becomes "further away" from the target distribution by adding in-domain data to the training set, overcoming this even greater degree of domain shift remains an open problem. For practitioners assembling their own datasets, a need for domain-level labels is a necessary precaution to take. But for practitioners using large datasets with an unknown number of domains, it's unclear how to overcome this effect without a priori knowledge of how many domains exist in a mixed-domain training set, as well as which classes correspond to which domains.


\newpage
\newpage
\bibliography{negative_transfer}
\bibliographystyle{icml2023}

\newpage
\appendix
\onecolumn

\section{Further Results}
Further results are for VLCS, PACS, and VisDA (where relevant), on ConvNext, ResNet18, ViT are shown in the following figures. 

\subsection{Proportion of Target Domain Data}

We plot the remainder of proportionality experiments in Figures 16 and 17. As a comparison point, we also plot the performance of the algorithm when it trains on no surrogate data, i.e. $\mathcal{T}_{in|train} \cup p*\mathcal{T}_{out|train}$. This allows us to determine during which regimes it would even be useful to have a surrogate dataset. We note that the performance gain of using data under domain shift becomes dramatically smaller with even very few examples from the target dataset. From these experiments, we determine that negative transfer still exists when $p$ is very small, but vanishes as $p$ gets larger. However, as $p$ gets larger, the improvement of the model due to the surrogate data also vanishes, and we find limited gains due to the auxiliary data.

\section{Further Details}

\subsection{Experimental Details}
 All PACS, VisDA, and VLCS data are resized to 224x224 and normalized with mean = $[0.485, 0.456, 0.406]$ and std=$[0.229, 0.224, 0.225]$. All models train with SGD and learning rate $0.001$ for five epochs with a standard cross-entropy loss, and we choose the model that performs the best on the validation set as the trained model. Note that the validation set is composed with the same in-class and out-class stratification that the training set is. For both of these datasets, we use batch sizes of 32 for the ResNet models, and batch sizes of 16 for the ConvNext and ViT-B models. All models start with pretrained Imagenet weights, and fine-tune on their respective training sets.


\begin{figure}
     \centering
     \begin{subfigure}[b]{0.45\textwidth}
         \centering
         \includegraphics[width=\textwidth]{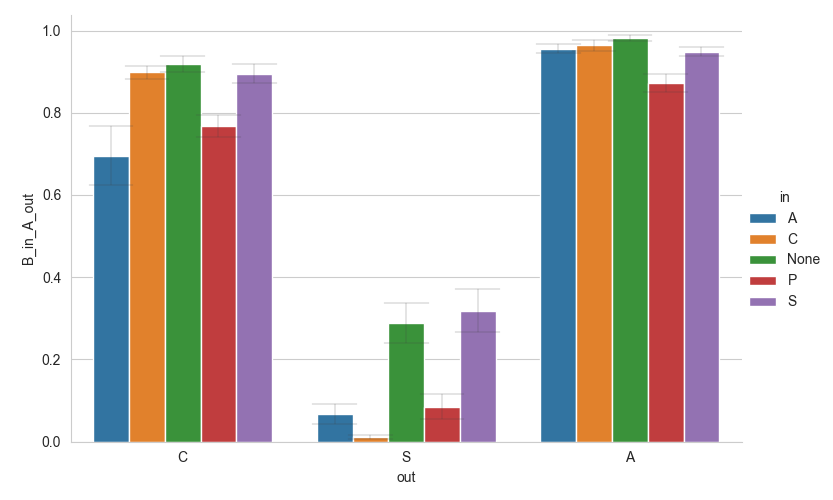}
         \caption{Target Domain P}
         \label{fig:pacs_s2_p_r18}
     \end{subfigure}
     \hfill
     \begin{subfigure}[b]{0.45\textwidth}
         \centering
         \includegraphics[width=\textwidth]{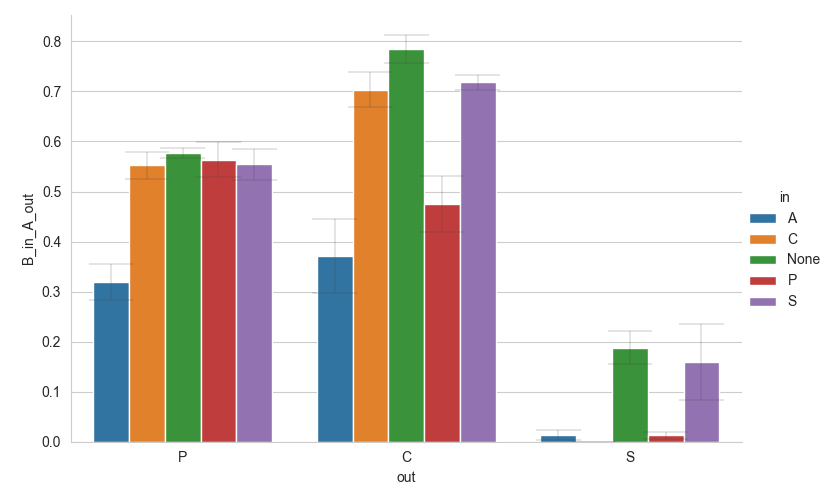}
         \caption{Target Domain A}
         \label{fig:pacs_s2_a_r18}
     \end{subfigure}
     \hfill
          \begin{subfigure}[b]{0.45\textwidth}
         \centering
         \includegraphics[width=\textwidth]{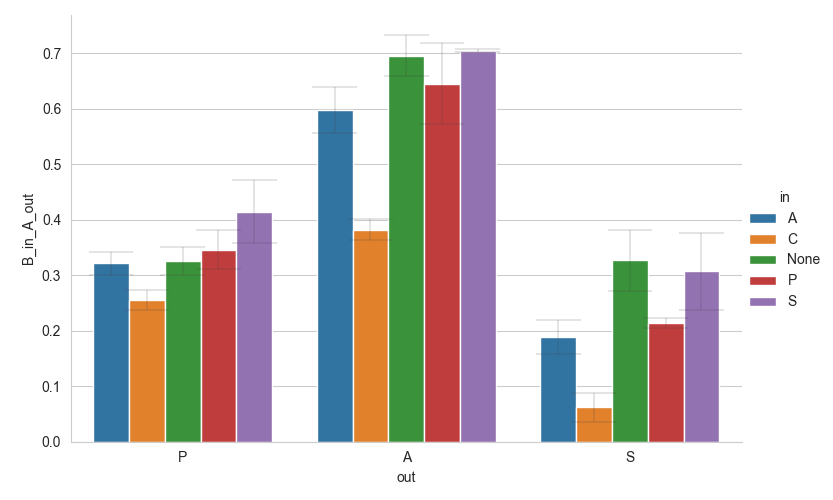}
         \caption{Target Domain C}
         \label{fig:pacs_s2_c_r18}
     \end{subfigure}
     \hfill
      \begin{subfigure}[b]{0.45\textwidth}
     \centering
     \includegraphics[width=\textwidth]{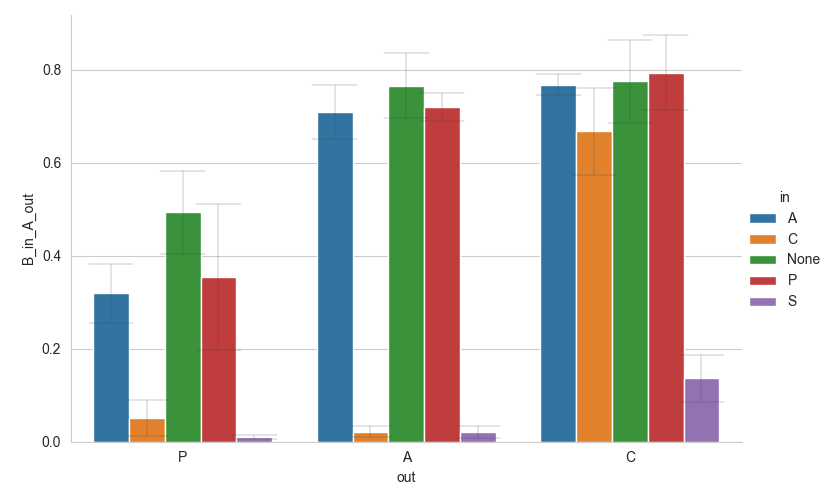}
     \caption{Target Domain S}
     \label{fig:pacs_s2_s_r18}
     \end{subfigure}
        \caption{PACS Two Sources experiments, resnet18}
        \label{fig:pacs_s2_r18}
\end{figure}

\begin{figure}
     \centering
     \begin{subfigure}[b]{0.45\textwidth}
         \centering
         \includegraphics[width=\textwidth]{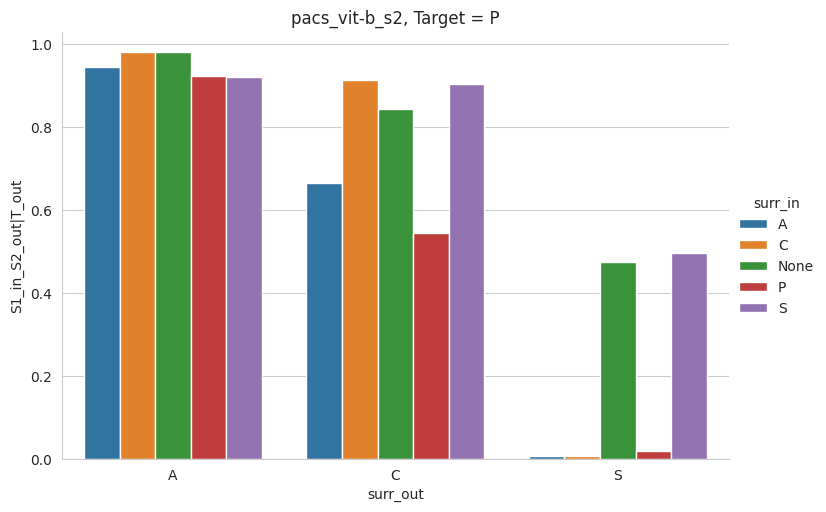}
         \caption{Target Domain P}
         \label{fig:pacs_s2_p_vit}
     \end{subfigure}
     \hfill
     \begin{subfigure}[b]{0.45\textwidth}
         \centering
         \includegraphics[width=\textwidth]{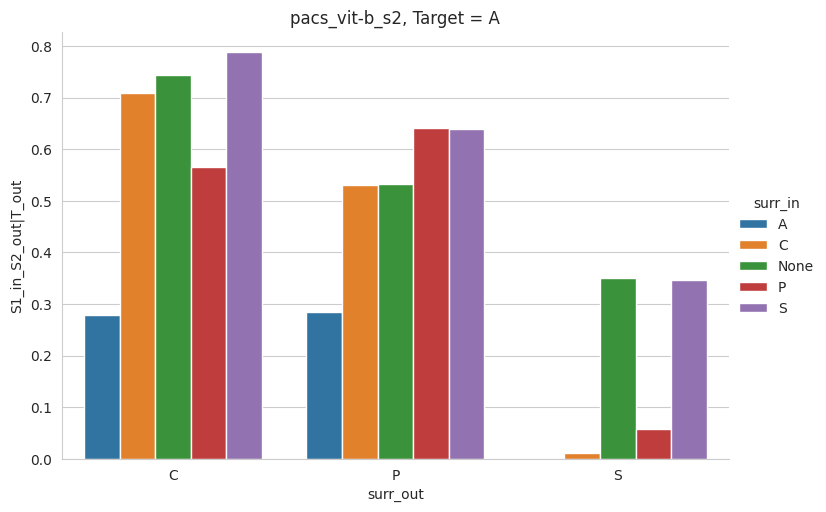}
         \caption{Target Domain A}
         \label{fig:pacs_s2_a_vit}
     \end{subfigure}
     \hfill
          \begin{subfigure}[b]{0.45\textwidth}
         \centering
         \includegraphics[width=\textwidth]{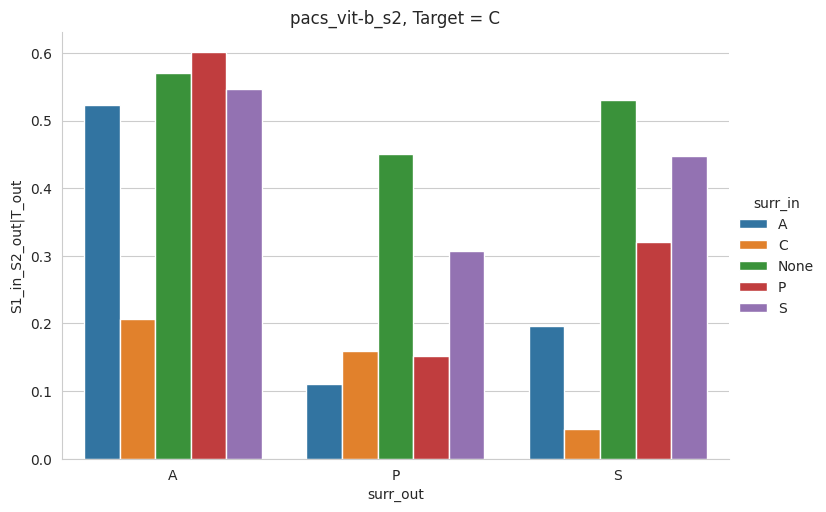}
         \caption{Target Domain C}
         \label{fig:pacs_s2_c_vit}
     \end{subfigure}
     \hfill
      \begin{subfigure}[b]{0.45\textwidth}
     \centering
     \includegraphics[width=\textwidth]{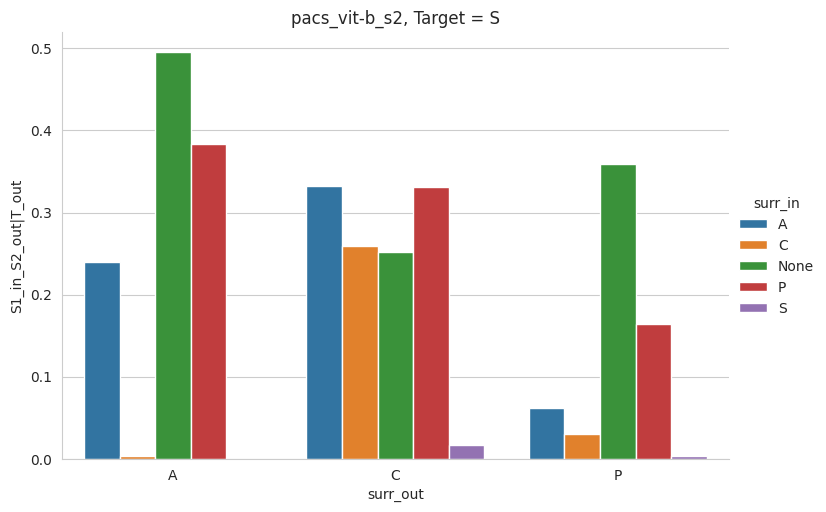}
     \caption{Target Domain S}
     \label{fig:pacs_s2_s_vit}
     \end{subfigure}
        \caption{PACS Two Sources experiments, ViT-B}
        \label{fig:pacs_s2_vit}
\end{figure}


\begin{figure}
     \centering
     \begin{subfigure}[b]{0.45\textwidth}
         \centering
         \includegraphics[width=\textwidth]{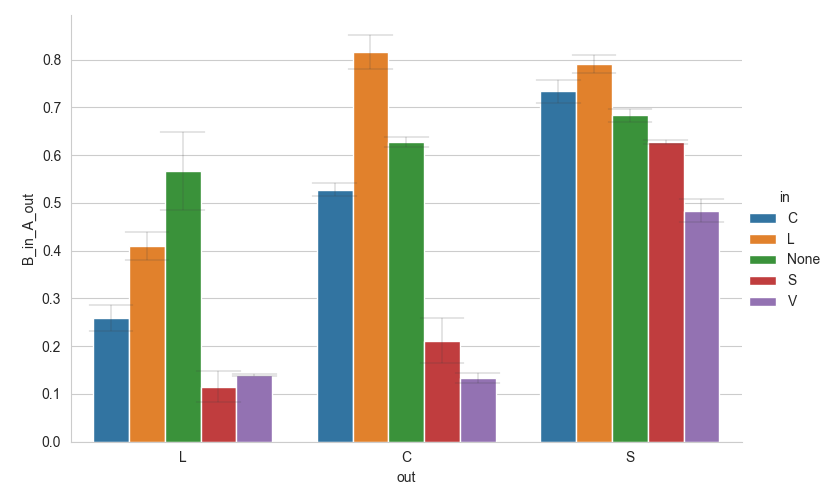}
         \caption{Target Domain V}
         \label{fig:vlcs_s2_v_r18}
     \end{subfigure}
     \hfill
     \begin{subfigure}[b]{0.45\textwidth}
         \centering
         \includegraphics[width=\textwidth]{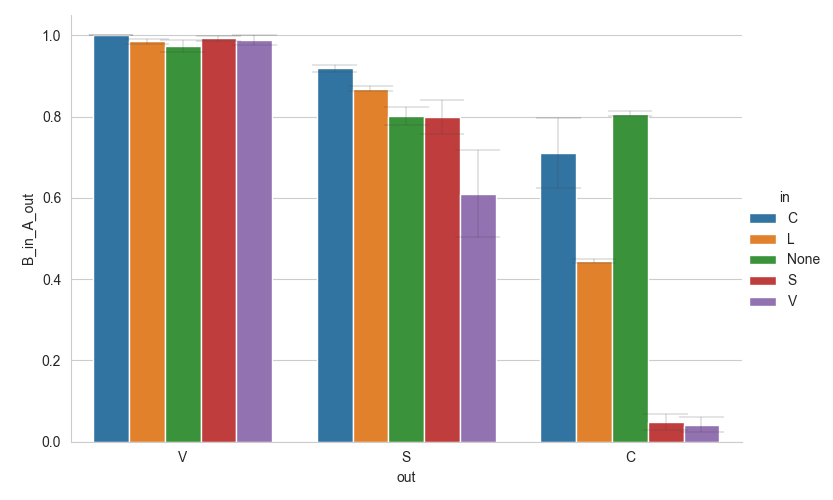}
         \caption{Target Domain L}
         \label{fig:vlcs_s2_l_r18}
     \end{subfigure}
     \hfill
          \begin{subfigure}[b]{0.45\textwidth}
         \centering
         \includegraphics[width=\textwidth]{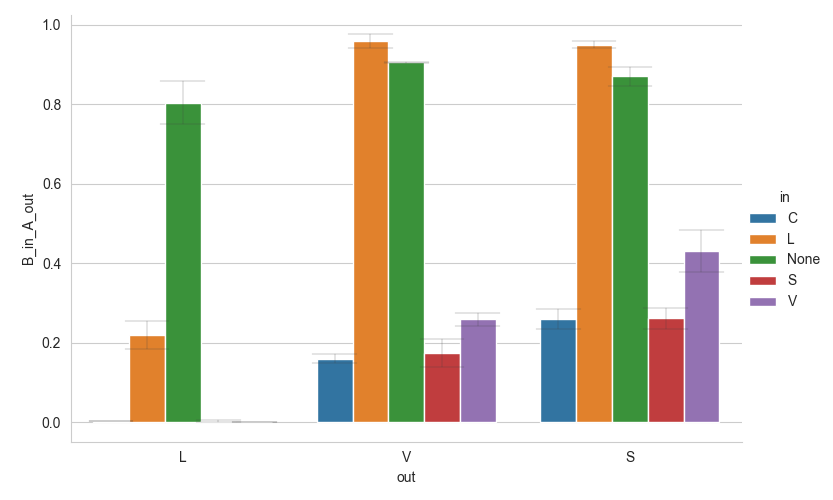}
         \caption{Target Domain C}
         \label{fig:vlcs_s2_c_r18}
     \end{subfigure}
     \hfill
      \begin{subfigure}[b]{0.45\textwidth}
     \centering
     \includegraphics[width=\textwidth]{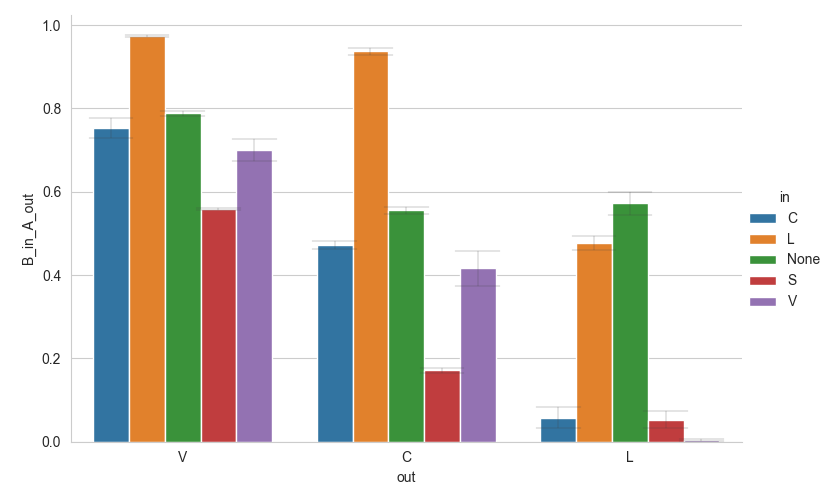}
     \caption{Target Domain S}
     \label{fig:vlcs_s2_s_r18}
     \end{subfigure}
        \caption{VLCS Two Sources experiments, ResNet18}
        \label{fig:vlcs_s2_r18}
\end{figure}

\begin{figure}
     \centering
     \begin{subfigure}[b]{0.45\textwidth}
         \centering
         \includegraphics[width=\textwidth]{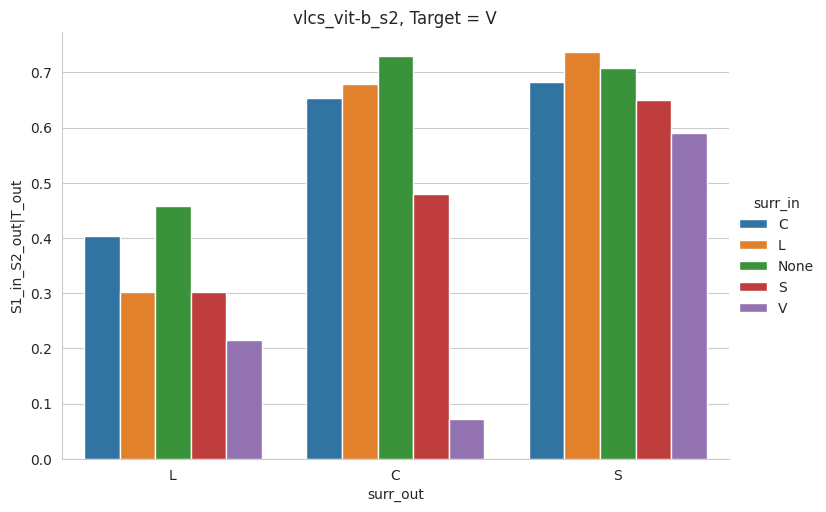}
         \caption{Target Domain V}
         \label{fig:vlcs_s2_v_vit}
     \end{subfigure}
     \hfill
     \begin{subfigure}[b]{0.45\textwidth}
         \centering
         \includegraphics[width=\textwidth]{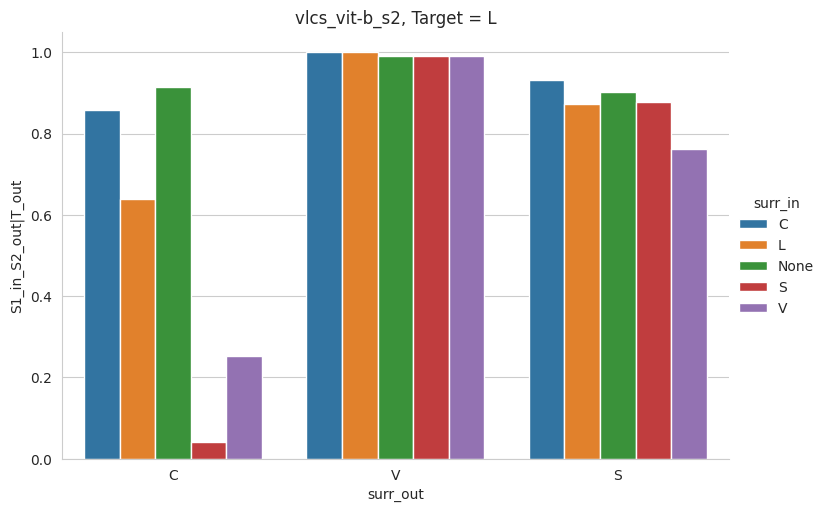}
         \caption{Target Domain L}
         \label{fig:vlcs_s2_l_vit}
     \end{subfigure}
     \hfill
          \begin{subfigure}[b]{0.45\textwidth}
         \centering
         \includegraphics[width=\textwidth]{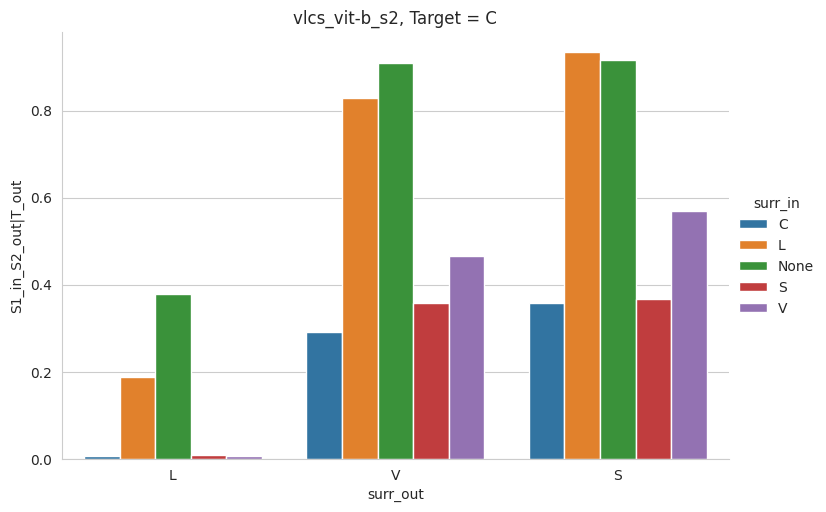}
         \caption{Target Domain C}
         \label{fig:vlcs_s2_c_vit}
     \end{subfigure}
     \hfill
      \begin{subfigure}[b]{0.45\textwidth}
     \centering
     \includegraphics[width=\textwidth]{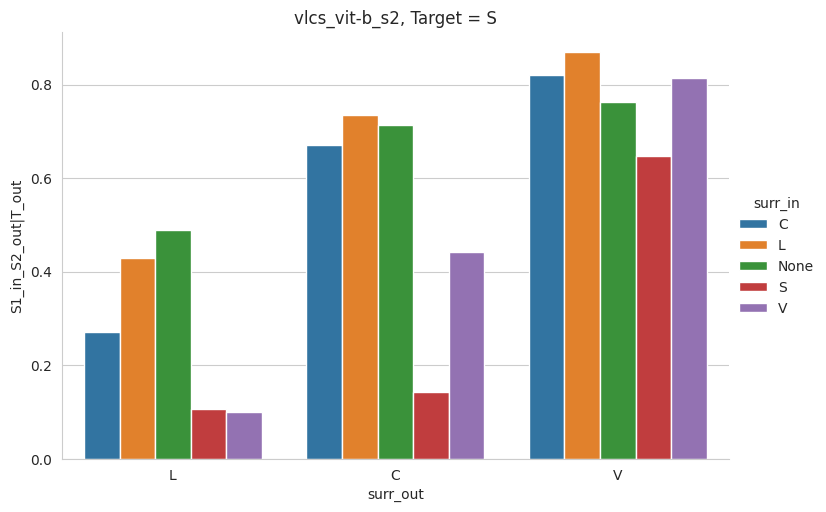}
     \caption{Target Domain S}
     \label{fig:vlcs_s2_s_vit}
     \end{subfigure}
        \caption{VLCS Two Sources experiments, ViT-B}
        \label{fig:vlcs_s2_vit}
\end{figure}



\begin{figure}
     \centering
     \begin{subfigure}[b]{.8\textwidth}
         \centering
         \includegraphics[width=\textwidth]{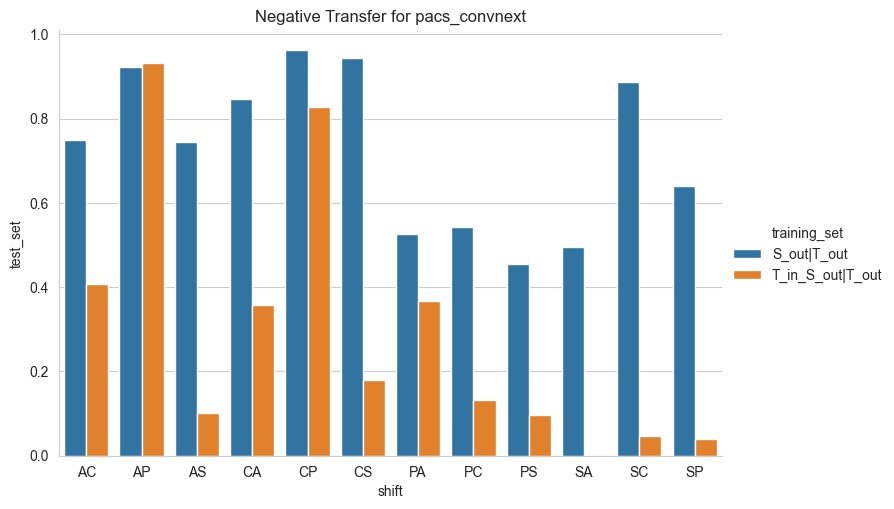}
         \caption{ConvNext}
         \label{fig:pacs_mixed_cn}
     \end{subfigure}
     \hfill
     \begin{subfigure}[b]{.8\textwidth}
         \centering
         \includegraphics[width=\textwidth]{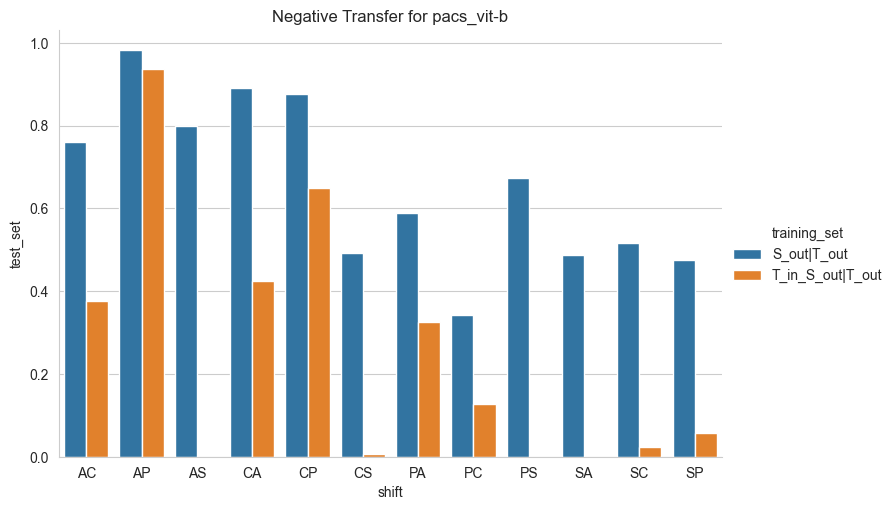}
         \caption{ViT-B}
         \label{fig:pacs_mixed_vit}
     \end{subfigure}
     \hfill
          \begin{subfigure}[b]{.8\textwidth}
         \centering
         \includegraphics[width=\textwidth]{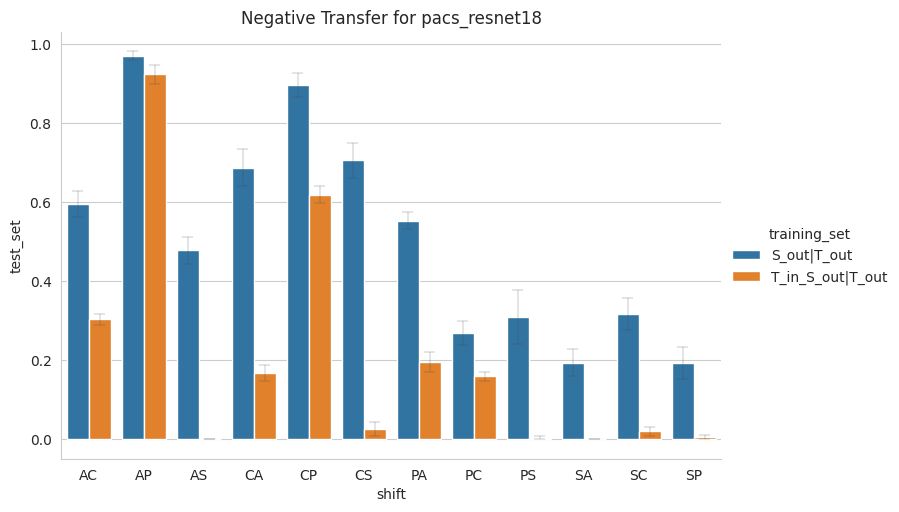}
         \caption{ResNet18}
         \label{fig:pacs_mixed_r18}
     \end{subfigure}
        \caption{PACS Negative Transfer Experiments}
        \label{fig:pacs_mixed}
\end{figure}

\begin{figure}
     \centering
     \begin{subfigure}[b]{.8\textwidth}
         \centering
         \includegraphics[width=\textwidth]{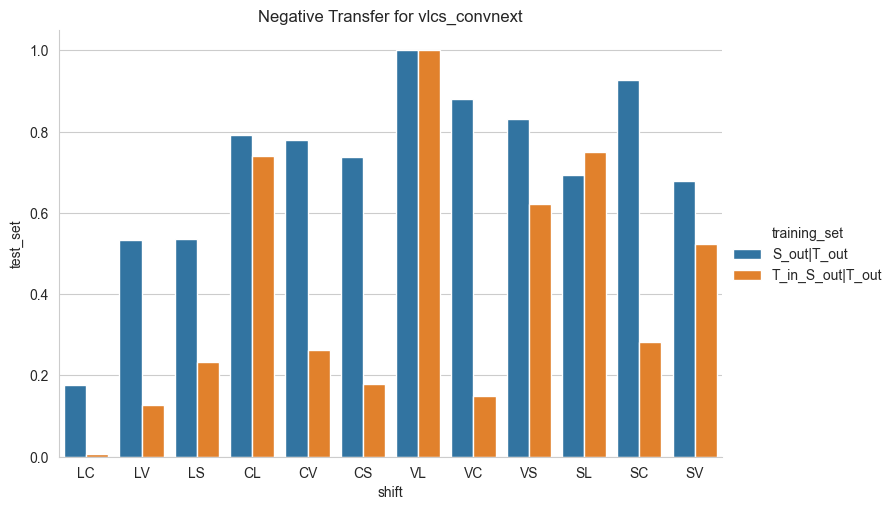}
         \caption{ConvNext}
         \label{fig:vlcs_mixed_cn}
     \end{subfigure}
     \hfill
     \begin{subfigure}[b]{.8\textwidth}
         \centering
         \includegraphics[width=\textwidth]{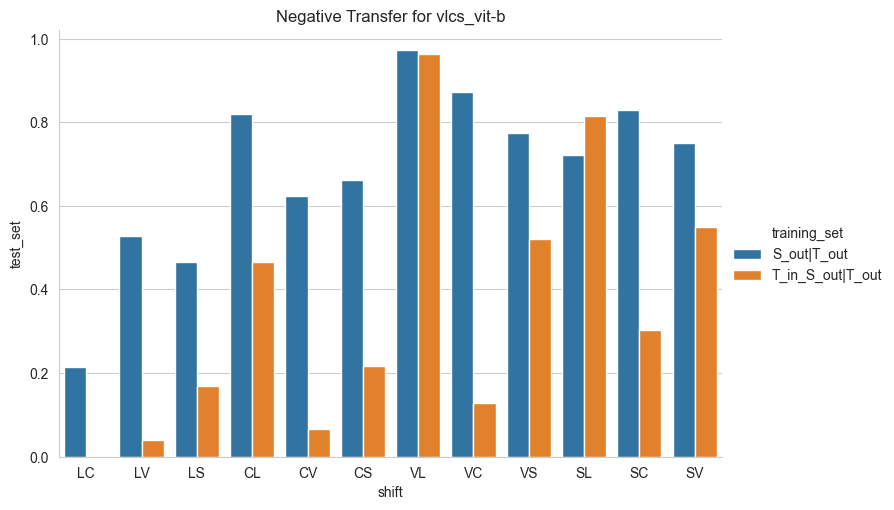}
         \caption{ViT-B}
         \label{fig:vlcs_mixed_vit}
     \end{subfigure}
     \hfill
          \begin{subfigure}[b]{.8\textwidth}
         \centering
         \includegraphics[width=\textwidth]{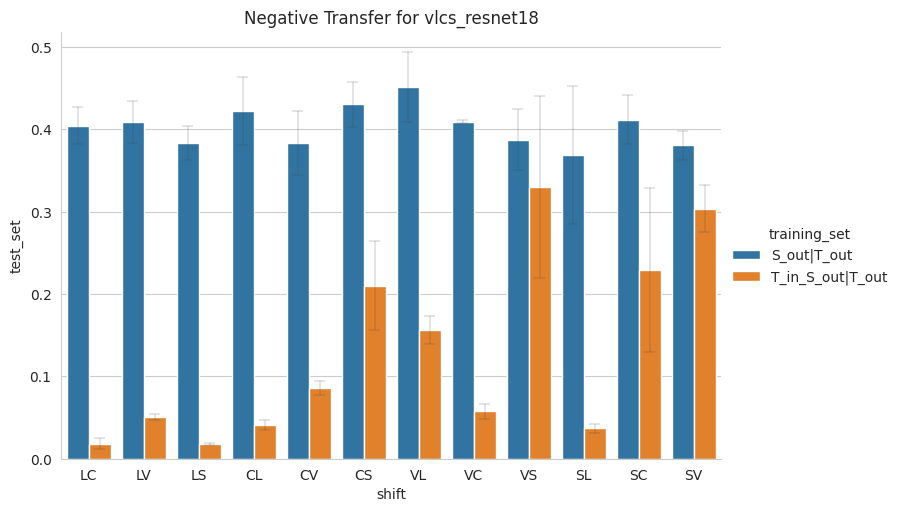}
         \caption{ResNet18}
         \label{fig:vlcs_mixed_r18}
     \end{subfigure}
        \caption{VLCS Negative Transfer Experiments}
        \label{fig:vlcs_mixed}
\end{figure}

\begin{figure}
     \centering
     \begin{subfigure}[b]{.48\textwidth}
         \centering
         \includegraphics[width=\textwidth]{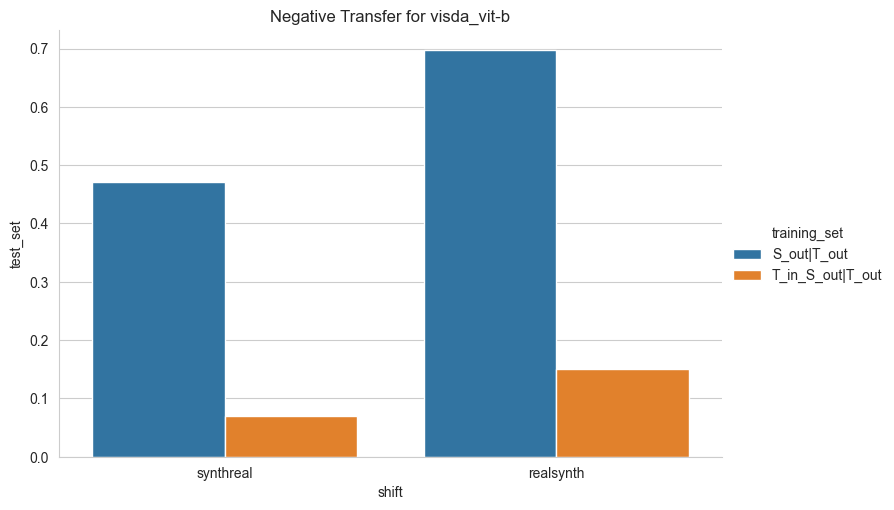}
         \caption{ViT-B}
         \label{fig:visda_mixed_vit}
     \end{subfigure}
     \hfill
          \begin{subfigure}[b]{.48\textwidth}
         \centering
         \includegraphics[width=\textwidth]{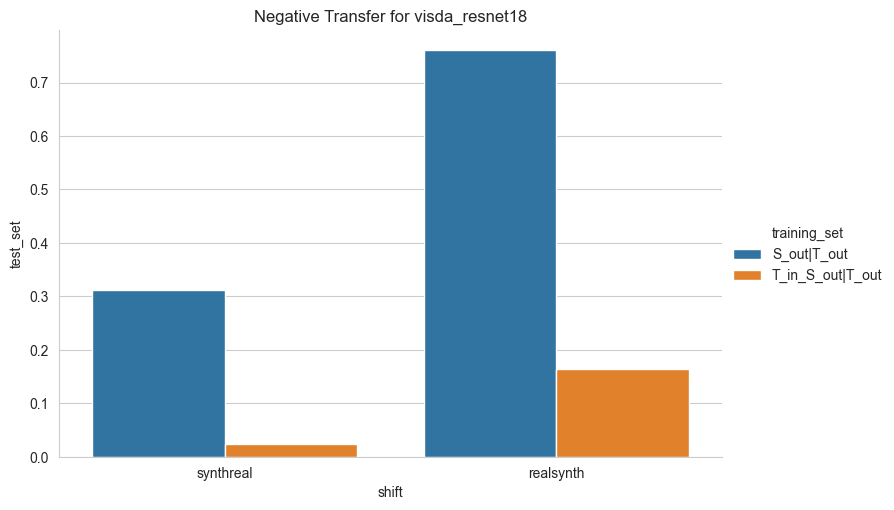}
         \caption{ResNet18}
         \label{fig:visda_mixed_r18}
     \end{subfigure}
        \caption{VisDA Negative Transfer Experiments}
        \label{fig:visda_mixed}
\end{figure}




 
\begin{figure}
     \centering
     \begin{subfigure}[b]{.3\textwidth}
         \centering
         \includegraphics[width=\textwidth]{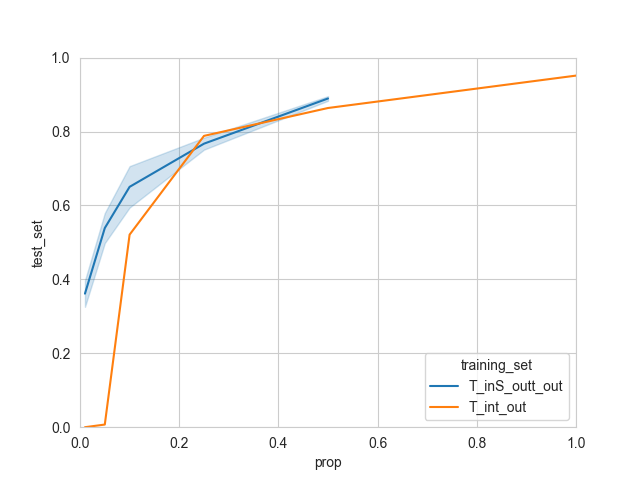}
         \caption{AC}
     \end{subfigure}
     \hfill
     \begin{subfigure}[b]{.3\textwidth}
         \centering
         \includegraphics[width=\textwidth]{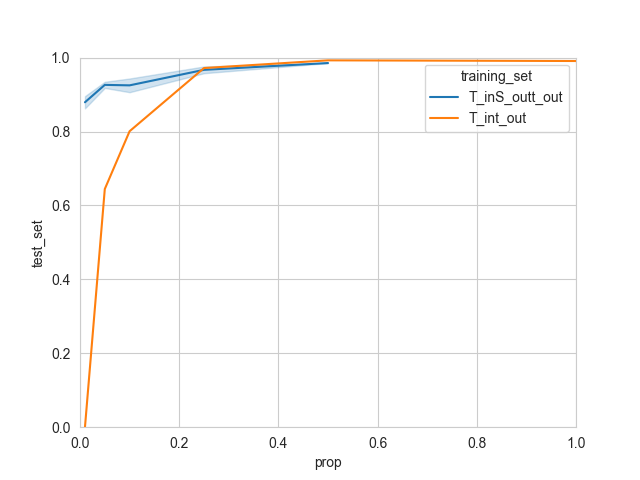}
         \caption{AP}
     \end{subfigure}
     \hfill
     \begin{subfigure}[b]{.3\textwidth}
         \centering
         \includegraphics[width=\textwidth]{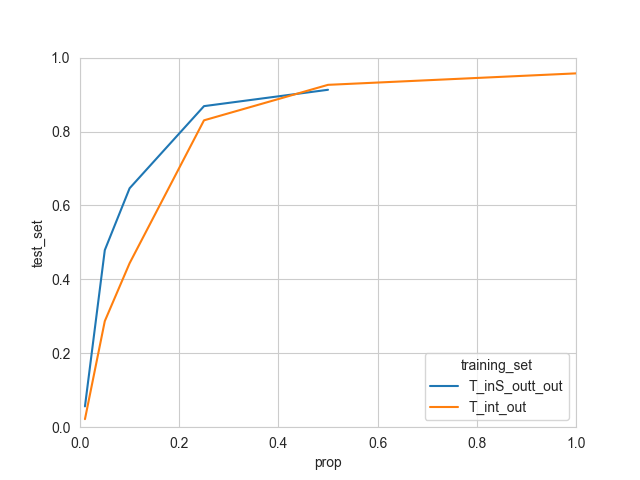}
         \caption{AS}
     \end{subfigure}
     \hfill
     \begin{subfigure}[b]{.3\textwidth}
         \centering
         \includegraphics[width=\textwidth]{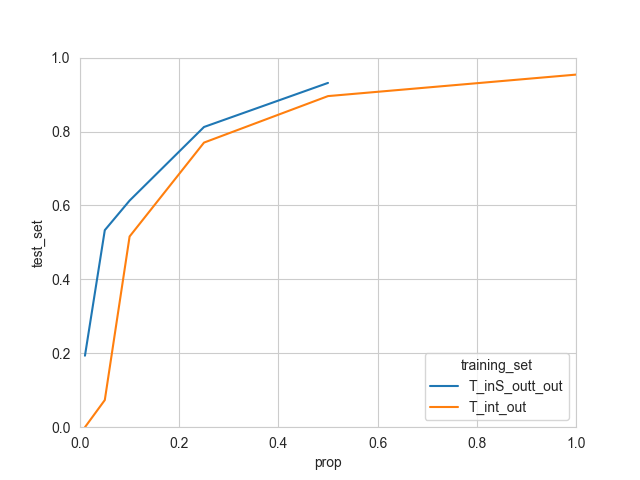}
         \caption{CA}
     \end{subfigure}
     \hfill
     \begin{subfigure}[b]{.3\textwidth}
         \centering
         \includegraphics[width=\textwidth]{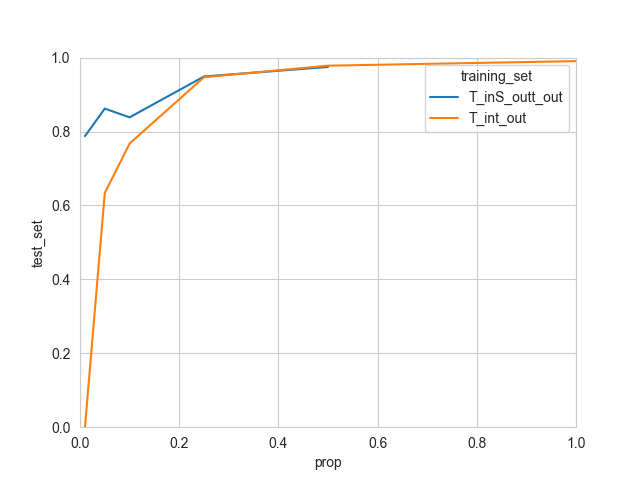}
         \caption{CP}
     \end{subfigure}
     \hfill
     \begin{subfigure}[b]{.3\textwidth}
         \centering
         \includegraphics[width=\textwidth]{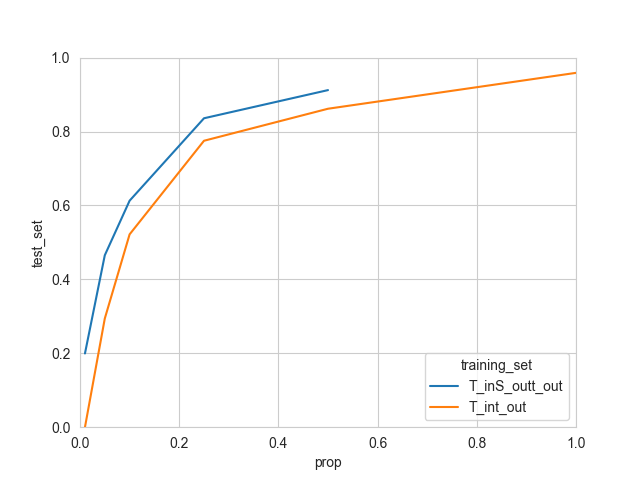}
         \caption{CS}
     \end{subfigure}
     \hfill
     \begin{subfigure}[b]{.3\textwidth}
         \centering
         \includegraphics[width=\textwidth]{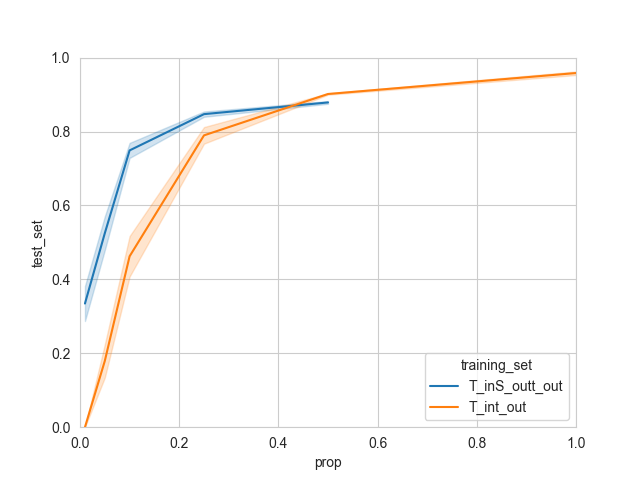}
         \caption{PA}
     \end{subfigure}
     \hfill
     \begin{subfigure}[b]{.3\textwidth}
         \centering
         \includegraphics[width=\textwidth]{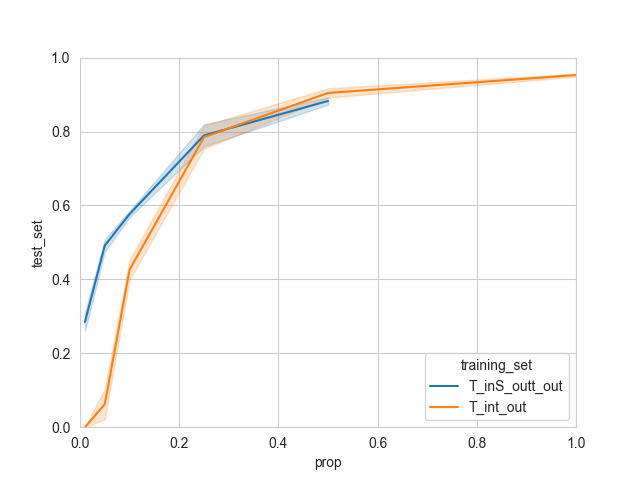}
         \caption{PC}
     \end{subfigure}
     \hfill
     \begin{subfigure}[b]{.3\textwidth}
         \centering
         \includegraphics[width=\textwidth]{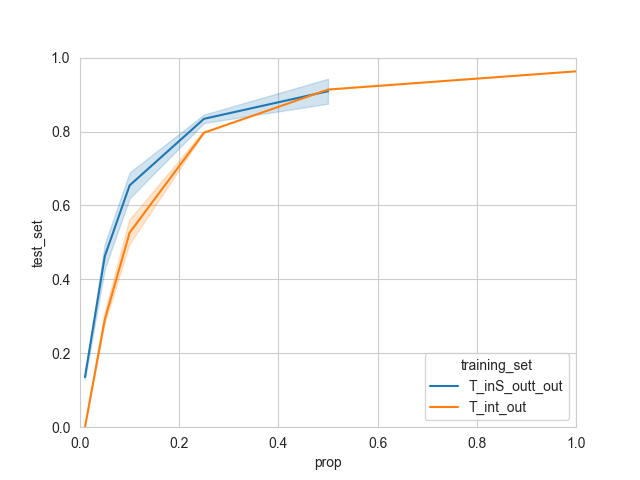}
         \caption{PS}
     \end{subfigure}
     \hfill
     \begin{subfigure}[b]{.3\textwidth}
         \centering
         \includegraphics[width=\textwidth]{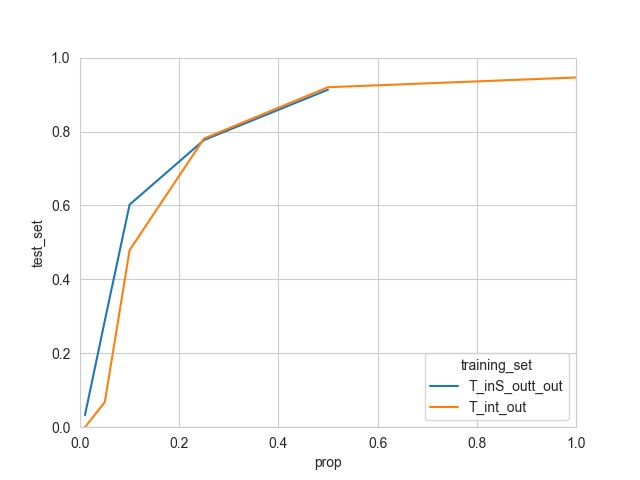}
         \caption{SA}
     \end{subfigure}
     \hfill
     \begin{subfigure}[b]{.3\textwidth}
         \centering
         \includegraphics[width=\textwidth]{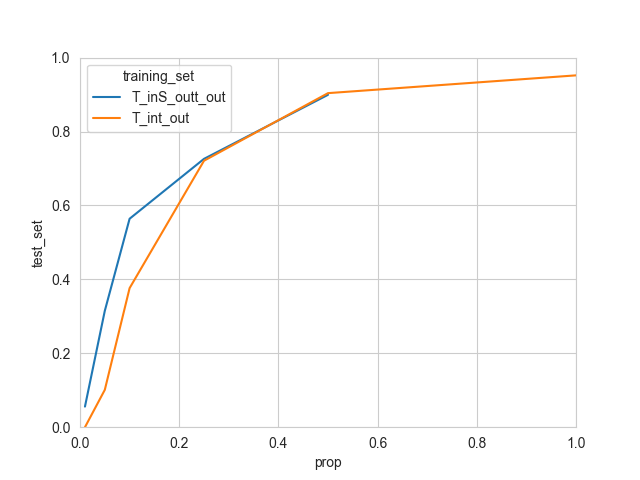}
         \caption{SC}
     \end{subfigure}
     \hfill
     \begin{subfigure}[b]{.3\textwidth}
         \centering
         \includegraphics[width=\textwidth]{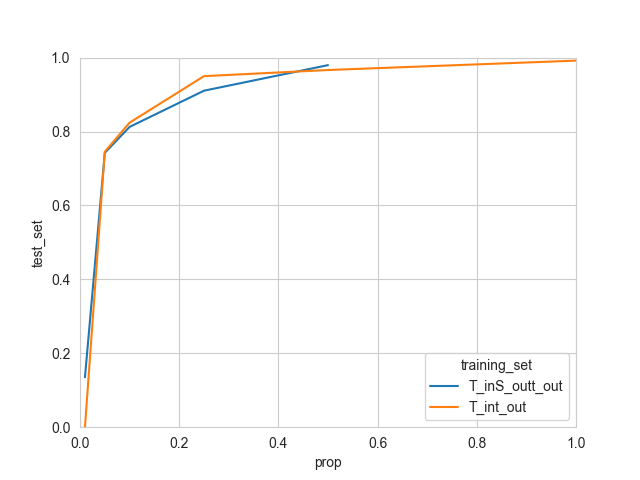}
         \caption{SP}
     \end{subfigure}
     \hfill
             \caption{PACS Proportionality Experiments, ResNet18. Figures are captioned as [Surrogate][Target]}
        \label{fig:pacs_prop_r18}
\end{figure}

\begin{figure}
     \centering
     \begin{subfigure}[b]{.3\textwidth}
         \centering
         \includegraphics[width=\textwidth]{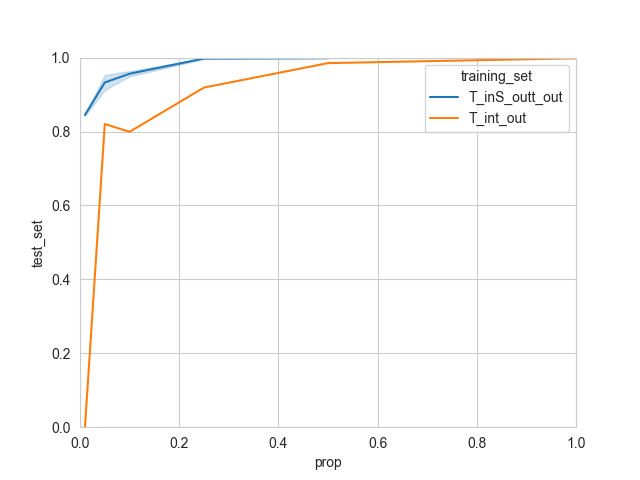}
         \caption{CL}
     \end{subfigure}
     \hfill
     \begin{subfigure}[b]{.3\textwidth}
         \centering
         \includegraphics[width=\textwidth]{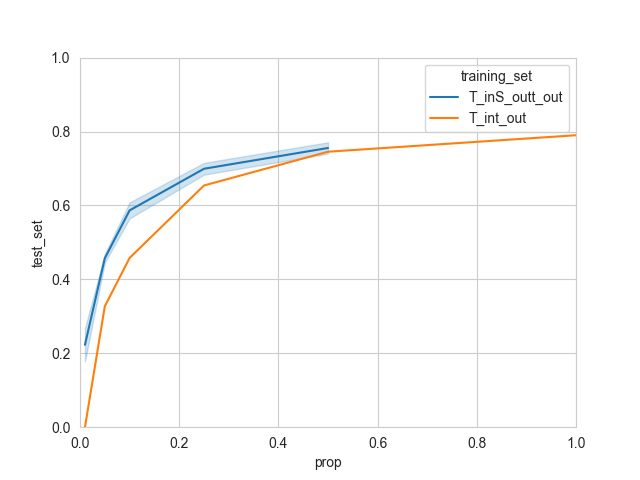}
         \caption{CS}
     \end{subfigure}
     \hfill
     \begin{subfigure}[b]{.3\textwidth}
         \centering
         \includegraphics[width=\textwidth]{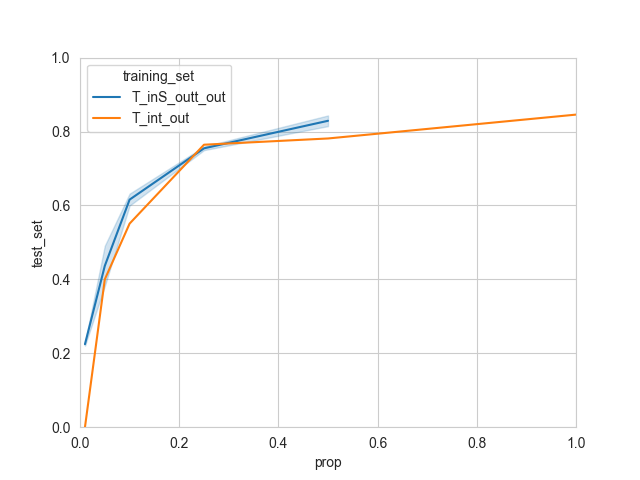}
         \caption{CV}
     \end{subfigure}
     \hfill
     \begin{subfigure}[b]{.3\textwidth}
         \centering
         \includegraphics[width=\textwidth]{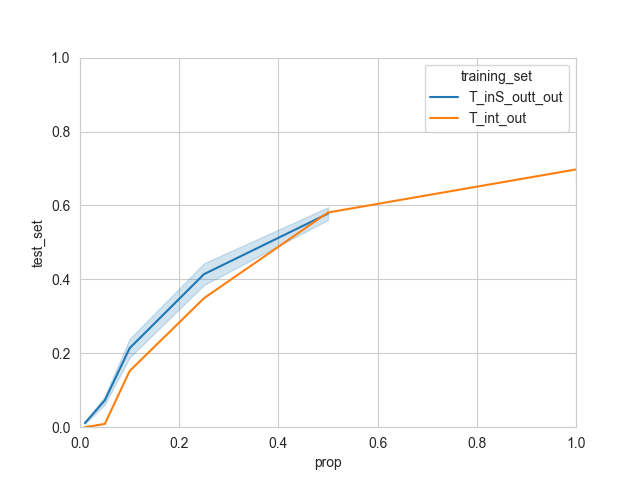}
         \caption{LC}
     \end{subfigure}
     \hfill
     \begin{subfigure}[b]{.3\textwidth}
         \centering
         \includegraphics[width=\textwidth]{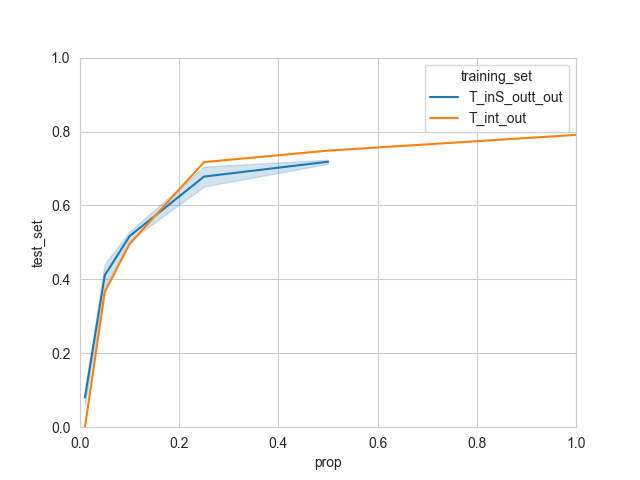}
         \caption{LS}
     \end{subfigure}
     \hfill
     \begin{subfigure}[b]{.3\textwidth}
         \centering
         \includegraphics[width=\textwidth]{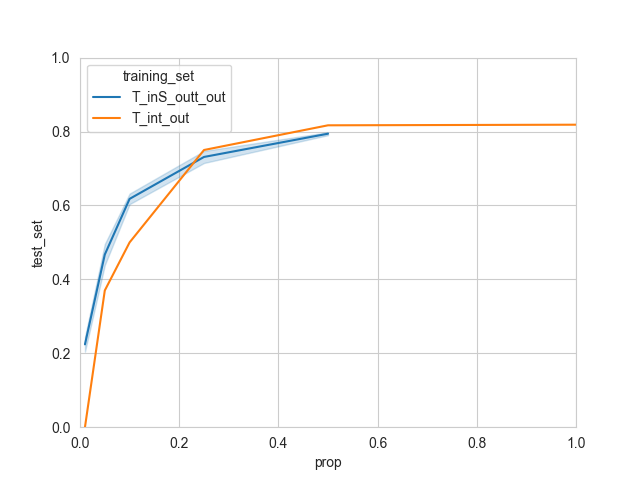}
         \caption{LV}
     \end{subfigure}
     \hfill
     \begin{subfigure}[b]{.3\textwidth}
         \centering
         \includegraphics[width=\textwidth]{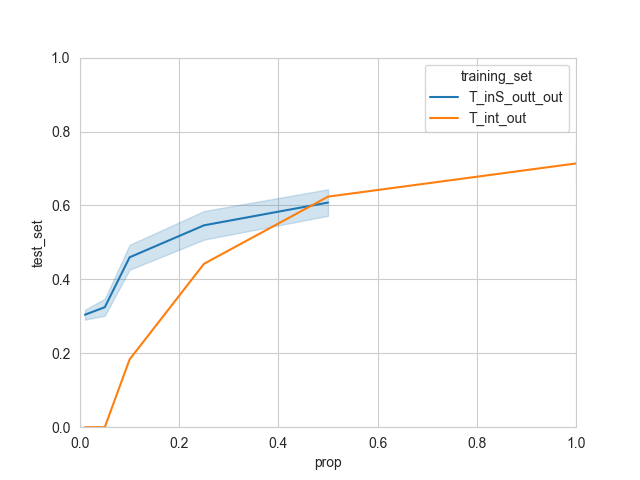}
         \caption{SC}
     \end{subfigure}
     \hfill
     \begin{subfigure}[b]{.3\textwidth}
         \centering
         \includegraphics[width=\textwidth]{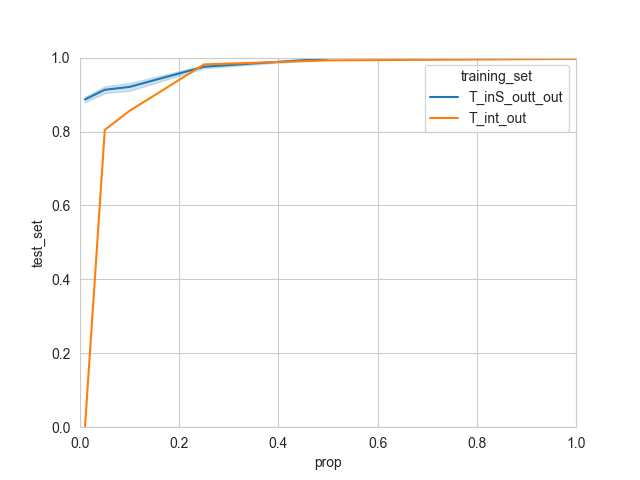}
         \caption{SL}
     \end{subfigure}
     \hfill
     \begin{subfigure}[b]{.3\textwidth}
         \centering
         \includegraphics[width=\textwidth]{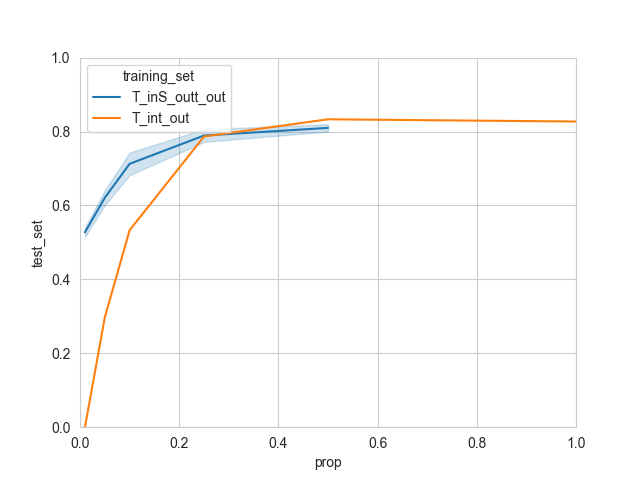}
         \caption{SV}
     \end{subfigure}
     \hfill
     \begin{subfigure}[b]{.3\textwidth}
         \centering
         \includegraphics[width=\textwidth]{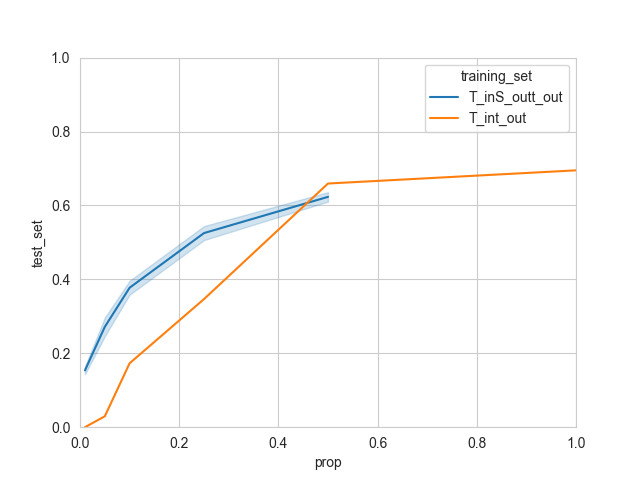}
         \caption{VC}
     \end{subfigure}
     \hfill
     \begin{subfigure}[b]{.3\textwidth}
         \centering
         \includegraphics[width=\textwidth]{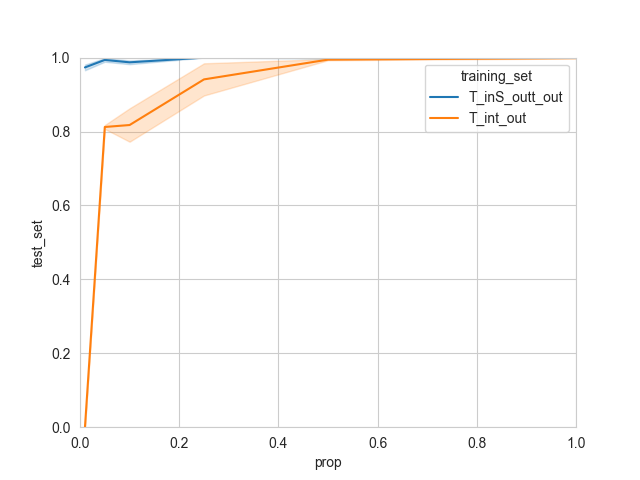}
         \caption{VL}
     \end{subfigure}
     \hfill
     \begin{subfigure}[b]{.3\textwidth}
         \centering
         \includegraphics[width=\textwidth]{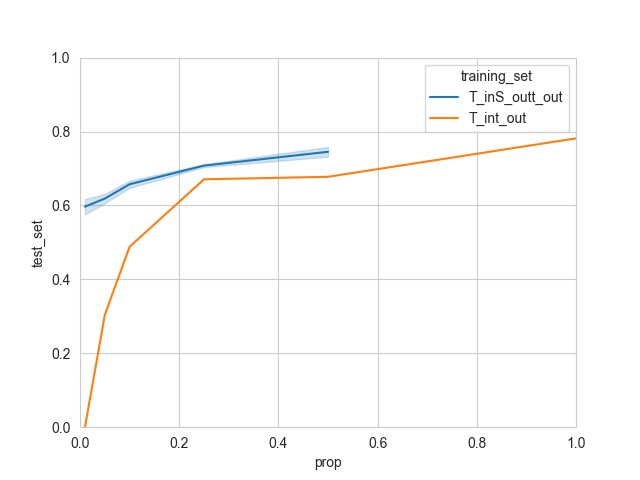}
         \caption{VS}
     \end{subfigure}
     \hfill
             \caption{VLCS Proportionality Experiments, ResNet18. Figures are captioned as [Surrogate][Target]}
        \label{fig:vlcs_prop_r18}
\end{figure}

\end{document}